%% file: rev-main.tex
\documentclass{article}
\usepackage[comma]{natbib}
\usepackage{url}
\usepackage{graphicx}
\usepackage{bm}
\usepackage{amsfonts}
\usepackage{amsmath}
\usepackage{algorithm}
\usepackage{amsthm}
\usepackage{xcolor}
\usepackage{algorithmic}
\usepackage{comment}
\usepackage{subcaption}
\usepackage{marginnote}
\usepackage{geometry}
\input{commands}

\begin{document}

\markboth{Meila \& Zhang}{Manifold Learning}

\title{Manifold learning: what, how, and why}

\author{Marina Meila,$^1$ Hanyu Zhang,$^2$}

\maketitle

\begin{abstract}
  \mmp{Abstract text, approximately 150 words.} Manifold learning (ML), known also as non-linear dimension reduction, is a set of methods to find the low dimensional structure of data. Dimension reduction \mmp{and describing low dim structure} for large, high dimensional data is not merely a way to reduce the data; the new representations and descriptors obtained by ML reveal the geometric shape of high dimensional point clouds, and allow one to visualize, denoise and interpret them. This survey presents the principles underlying ML, the representative methods, as well as their statistical foundations from a practicing statistician's perspective. It describes the trade-offs, and what theory tells us about the parameter and algorithmic choices we make in order to obtain reliable conclusions. 
\end{abstract}
\tableofcontents

\section{Introduction}
\label{sec:intro}
\input{rev-intro}

\section{Mathematical background}
\label{sec:bg}
\input{rev-bg}

\section{Premises and paradigms in manifold learning}
\label{sec:ml}
\input{rev-ml}

\section{Embedding algorithms}
\label{sec:algs}
\input{rev-algs}
\input{rev-ies} 

\section{Statistical basis of manifold learning}
\label{sec:geo}
\input{rev-geo}

\section{Applications of manifold learning}
\label{sec:appli}
\input{rev-appli}

\section{Conclusion}
\label{sec:conclusion}
\input{rev-conclusion}

\bibliographystyle{ar-style1}
\bibliography{refs}

\end{document}

%% file: commands.tex
\newenvironment{itemize*}{
\begin{itemize}
\setlength{\parskip}{0em}
\setlength{\topparskip}{0em}
}
{\end{itemize}}

\newenvironment{enumerate*}{
\begin{enumerate}
\setlength{\parskip}{0em}
\setlength{\topparskip}{0em}
}
{\end{enumerate}}

\newcommand{\mmp}[1]{}
\newcommand{\hanyuz}[1]{}

\newcommand{\myemph}[1]{{\em #1}}
\newcommand{\mydef}[1]{\emph{#1}}

\newcommand{\lapalg}{{\sc Laplacian}}

\newcommand{\LEalg}{{\sc LE}}    

\newcommand{\dmalg}{{\sc DM}} 
\newcommand{\ltsaalg}{{\sc LTSA}}
\newcommand{\isomapalg}{{\sc Isomap}}
  
\newcommand{\rralg}{{\sc RiemannianRelaxation}}
\newcommand{\tsnealg}{{\sc t-SNE}}
\newcommand{\umapalg}{{\sc UMAP}}
\newcommand{\forcealg}{{\sc ForceAtlas}}

\newcommand{\beq}{\begin{equation}}
\newcommand{\eeq}{\end{equation}}
\newcommand{\beqa}{\begin{eqnarray}}
\newcommand{\eeqa}{\end{eqnarray}}
\newcommand{\beqas}{\begin{eqnarray*}}
\newcommand{\eeqas}{\end{eqnarray*}}

\newcommand{\bit}{\begin{itemize}}
\newcommand{\eit}{\end{itemize}}
\newcommand{\bits}{\begin{itemize*}}
\newcommand{\eits}{\end{itemize*}}
\newcommand{\benum}{\begin{enumerate}}
\newcommand{\eenum}{\end{enumerate}}
\newcommand{\benums}{\begin{enumerate*}}
\newcommand{\eenums}{\end{enumerate*}}

\newcommand{\fracpartial}[2]{\frac{\partial #1}{\partial  #2}}

\newcommand{\norm}[1]{\lvert\lvert{#1}\lvert\lvert}

\newcommand{\dataset}{{\cal D}}
\newcommand{\onevector}{{\mathbf 1}}
\newcommand{\rrr}{{\mathbb R}}
\newcommand{\dddF}{\operatorname{d}\!F}
\newcommand{\dist}{\operatorname{dist}}

\newcommand{\M}{{\cal M}}
\newcommand{\N}{{\cal N}}
\newcommand{\ponM}{\bm{p}}

\newcommand{\tgvec}{\bm{v}}
\newcommand{\T}{{\cal T}}
\newcommand{\tp}{\T_{\bm{p}}\mathcal{M}}
\newcommand{\F}{{\cal F}}
\newcommand{\neigh}{{\cal N}}
\newcommand{\loss}{{\mathcal L}}

\newcommand{\epps}{h}  

\newcommand{\Ab}{{\mathbf A}}
\newcommand{\Kb}{{\mathbf K}}

\newcommand{\Xb}{{\mathbf X}}
\newcommand{\Yb}{{\mathbf Y}}
\newcommand{\Vb}{{\mathbf V}}
\newcommand{\Wb}{{\mathbf W}}
\newcommand{\Db}{{\mathbf D}}

\newcommand{\pb}{\bm{p}} 
\newcommand{\ub}{\bm{u}} 
\newcommand{\vb}{\bm{v}} 
\newcommand{\xb}{\bm{x}}
\newcommand{\yb}{\bm{y}}

\newcommand{\bw}{h}  
\newcommand{\Lbw}{\mathbf{L}}  

\newcommand{\g}{\mathbf{g}} 
\newcommand{\gpush}{\tilde{\mathbf{g}}} 
\newcommand{\ginv}{\tilde{\mathbf{h}}} 

\newcommand{\hatdim}{\hat{d}}

\newcommand{\rank}{\operatorname{rank}}

\newcommand{\diag}{\operatorname{diag}}

\theoremstyle{plain}

\theoremstyle{definition}

\theoremstyle{remark}

%% file: rev-intro.tex
Modern data analysis tasks often face challenges of high dimension and thus nonlinear dimension reduction techniques emerge as a way to construct maps from high dimensional data to their corresponding low dimensional representations.  Finding such low dimensional representations of high dimensional data is beneficial in several aspects. This saves space and processing time. More importantly, the low dimensional representation often provides a better understanding of the \emph{intrinsic} structure of data, which often leads to better features that can be fed into further data analysis algorithms. This survey paper reviews the mathematical background, methodology, and recent development of nonlinear dimension reduction techniques. These techniques have been developed for two decades since two seminal works: \citet{tenenbaum00:isomap} and \citet{roweis:00}, and are widely used in various data analysis jobs, especially in scientific research.
%

Before nonlinear dimension reduction emerged, Principal Component
Analysis (PCA) was already widely accepted \citep{PCA}. Intuitively, PCA assumes that high dimensional data living in $\rrr^D$ 
lie around a lower dimensional linear subspace of $\rrr^D$ and seeks to
find the best linear subspace such that data points projected onto
this subspace have minimal reconstruction error. Nonlinear dimension reduction algorithms extend this idea
by assuming data are supported on smooth nonlinear low dimensional
geometric objects, i.e., manifolds embedded in $\rrr^D$, and find maps
that send the samples into lower dimensional coordinates while preserving some intrinsic geometric information.

In this survey, we start with a brief introduction to the main
differential geometric concepts underlying ML, elaborating on the
geometric information that manifolds carry
(Section \ref{sec:bg}). Then, in Section \ref{sec:ml}, we describe
the paradigm of manifold learning, with three possible sub-paradigms,
each producing a different representation of the data manifold. The
rest of the paper focuses on one of these, namely on the so-called
embedding algorithms. In Section \ref{sec:algs}, we survey representative manifold learning algorithms and their variants.
We also discuss the parameter choices,  as well as some pitfalls, which leads
to the discussion in Section \ref{sec:geo}, where we present the
statistical aspects and statistical results supporting these
methods. This section also includes 
 the estimation of crucial manifold descriptors from data: the Laplace-Beltrami
operator, Riemannian metrics, tangent space, intrinsic dimensions.
Section \ref{sec:appli} discusses applications, connecting with related statistics problems, and  Section \ref{sec:conclusion} concludes the
survey.

\mmp{some thoughts here:ML is estimation of an unknown (smooth) manifold $\M$ from data, or of some descriptor of $\M$ such as dimension, curvature, tangent at a point.}

%% file: rev-bg.tex
\subsection{Notations}
\label{sec:notation}
In this survey, we use the notation $\mathbb{R}^D$ to represent the
$D$-dimensional Euclidean space. A manifold is denoted as
$\M$. Lowercase Greek letters such as $\varphi, \phi, \psi, \rho,
\cdots$ represent functions mapping from $\M$ to a subset of Euclidean
space, while English letters like $f, g, h, \cdots$ denote functions
between real spaces. The notation $C^\ell$ refers to the class of
functions with continuous derivatives up to order $\ell$, and
$C^\infty$ represents the class of indefinitely differentiable
functions. The Kronecker delta is symbolized by
$\bm{\delta}_{ij}$.Bold lowercase English letters, like $\bm{v}$ and
$\bm{x}$, are used to denote vectors in Euclidean space. A dataset
containing $n$ data points is represented as the set $\dataset =
\{\bm{x}_i\}_{i=1}^n$. Bold uppercase letters, such as $\mathbf{A, B,
  C}$, denote matrices. By convention, we treat a single data point as
a column vector and use the matrix $\mathbf{X} \in \mathbb{R}^{n\times
  D}$ to represent the data matrix of a dataset with $n$ data points,
each being a vector in $D$-dimensional Euclidean space, while
$\Yb\in\rrr^{n\times m}$ will represent the same data mapped into $m$
dimension by a manifold learning algorithm. The notation $\lVert
\bm{v} \lVert$ signifies the $\ell_2$ norm of vector $\bm{v}$.\mmp{, while
$\lVert \mathbf{A} \lVert$ denotes the spectral norm of matrix
$\mathbf{A}$. The Frobenius norm of matrix $\mathbf{A}$ is
  represented as $\lVert \mathbf{A} \lVert_F$. We can also remove
  the spectral norm?} Throughout this survey, we also assume that all
functions are {\em smooth}, i.e., continuously differentiable as many
times as necessary.

\subsection{Manifold and Embedding}
\label{sec:mani-defi}
\paragraph*{Manifolds and coordinate charts}
Readers are referred to \citet{smoothmfd,DoCarmo} for a rigorous introduction to manifolds and differential geometry. Intuitively, the notion of a manifold is a generalization of curves and surfaces. 
\mmp{encapsulate this in a fig?/margin?}
\hanyuz{Move this to introduction?
The example in \ref{} shows that the rotation of an ethanol molecule has the structure of a torus $\mathbb{T}^2$. When parametrized by $\theta,\phi \in [0.2\pi)$, a torus in $\mathbb{R}^3$ can be represented by
\begin{equation}
    x(\theta,\phi)=(R+r\cos\theta)\cos\phi,\quad y(\theta,\phi)=(R+r\cos\theta)\sin\phi,\quad z(\theta,\phi)=r\sin\phi\;.
    \label{eq:torusdefi}
\end{equation}}
Mathematically, $\M$ is a \textbf{smooth manifold} of dimension $d$ (also called a $d$-manifold) if it is a topological space such that
\begin{itemize}
    \item For each $\bm{p}\in\M$, there exists a mapping $\varphi$ and an open neighborhood $U\subset\M$ of $\bm{p}$ such that $\varphi:U\rightarrow \varphi(U)$ is bijective and both $\varphi,\varphi^{-1}$ are smooth. Such pair $(U,\varphi)$ is called a \textbf{chart}, and $\varphi^{-1}:\rrr^d\rightarrow \M$ is called \textbf{local coordinate}. Hence, local coordinates map tuples in $\rrr^d$ to points on $\M$. 
    \item For any two points $\bm{p},\bm{p}'\in\M$ and two charts $(U,\varphi),(V,\phi)$ containing them, if $U\cap V\neq \emptyset$, then map $\varphi\circ\phi^{-1}$ is smooth on $\phi(U\cap V)$ and has a smooth inverse.
\end{itemize}
Hence, a smooth manifold is a set that locally around every point
resembles an open set in Euclidean space and for which transitions between charts are seamless. Moreover, with the help of smooth coordinate charts, one can define differentiable functions of a manifold, or between two manifolds, in a natural way.  
\begin{marginnote}
Technically, $\M$ is also required to be Hausdorff and second-countable. But most objects statisticians work with satisfy these conditions. Details about this mathematical definition of smooth manifolds can be found in the differential geometry textbook, such as \citet{DoCarmo}.
\end{marginnote}

\begin{figure}
\centering
\begin{subfigure}{0.45\textwidth}
\includegraphics[width=0.8\textwidth]{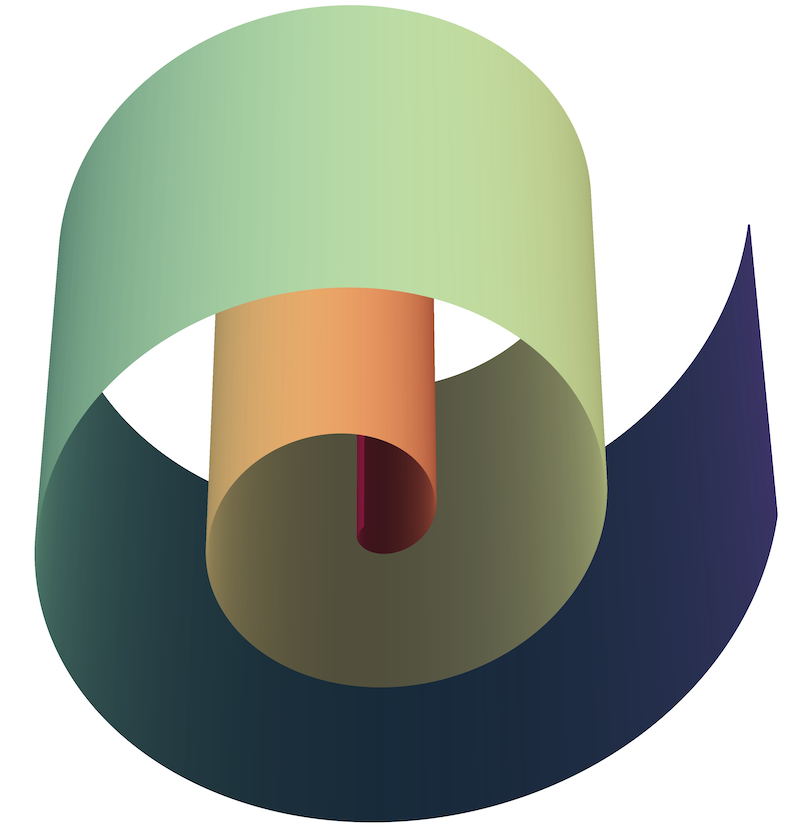}
\caption{Swiss Roll\label{fig:swissroll}}
\end{subfigure}
\begin{subfigure}{0.45\textwidth}
\includegraphics[width=0.8\textwidth]{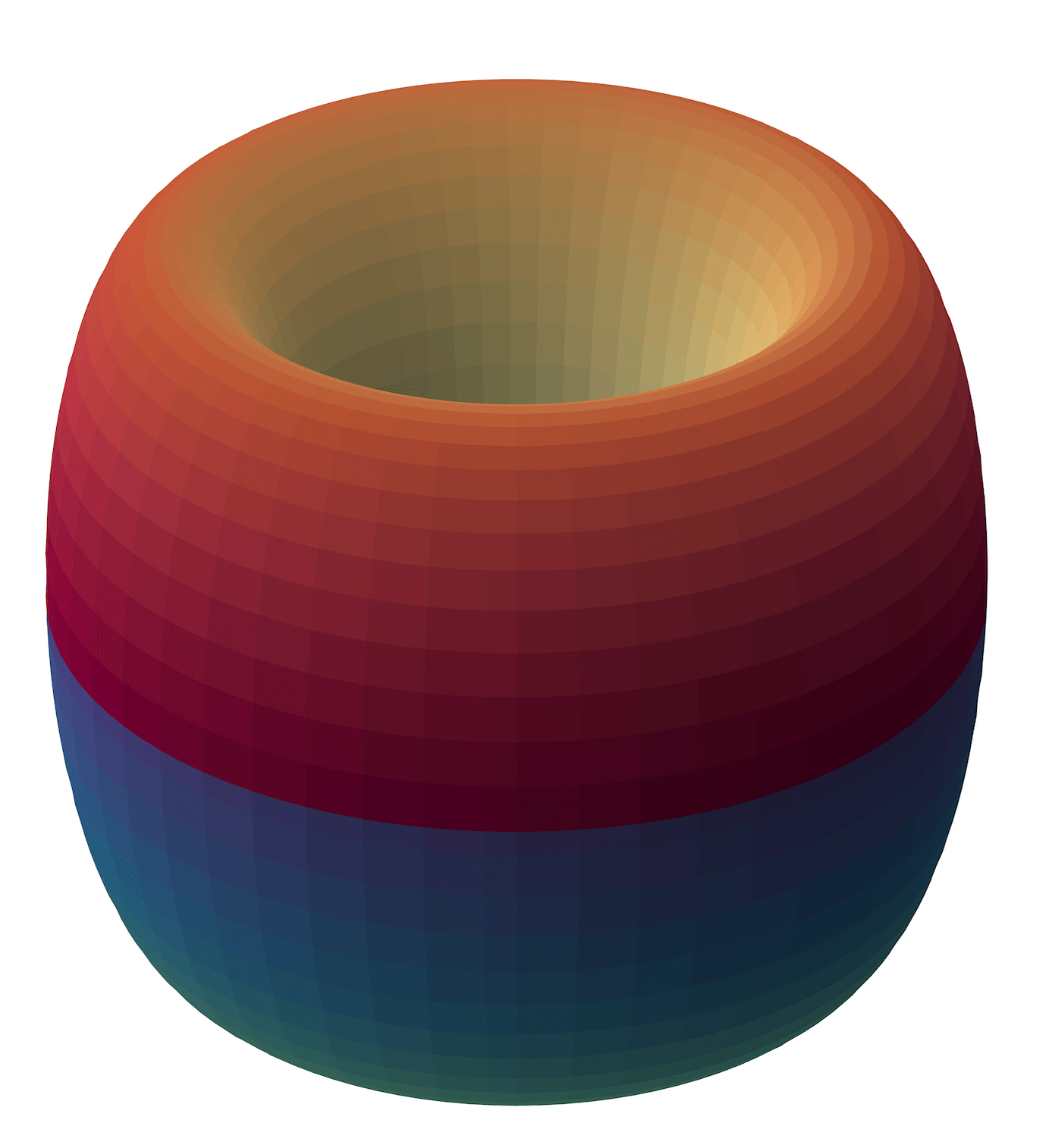}
\caption{Torus\label{fig:torus}}
\end{subfigure}
\end{figure}

The simplest example of a manifold is $\rrr^d$ itself, which has a
single, global coordinate chart. The ``swiss roll'' in Figure
\ref{fig:swissroll} is 2-manifolds
that also admits a global coordinate chart (into $\rrr^2$). A sphere,
or the torus in Figure \ref{fig:torus} are also 2-manifolds, but they
cannot be covered by a single chart (they each require at least two),
as cartographers well know. 

Note also that coordinate charts are not unique; $(U,\tilde{\phi})$
with $\tilde{\phi}=\tau\circ\phi$ is also a coordinate chart whenever
$\tau:\phi(U)\rightarrow \rrr^d$ is smoothly invertible ($\tau$ in this case is a change of variables). While the multiplicity of charts, atlases and
coordinate functions can be daunting at first sight, the framework of
differential geometry is set up so that calculus, geometric, and
topological quantities related to a manifold $\M$ are independent of
the coordinates chosen. For example, by compatibility, it follows that
the dimension $d$ must be the same for all charts and atlases. Hence,
$d$ is called the {\em intrinsic dimension} of the manifold $\M$. 

For a data scientist, the above means that, (1),
they can work in the coordinate system of their choice, and intrinsic
quantities like $d$ will remain invariant. But, (2),
care must be taken when the low dimensional data from two different
algorithms, or from different samples are being compared, because
these may not be in the same coordinate system.

\paragraph*{ Embeddings}
In differential geometry, an {\em embedding} is a smooth map between two manifolds $F:\M\rightarrow \N$ whose inverse $F^{-1}:\F(\M)\subset\N\rightarrow \M$ exists and is also smooth.
Of special interest is the case $\M\subset \N$; then $\M$ is said to be a submanifold of $\N$. Commonly in statistics, the high dimensional data lie originally in $\N=\rrr^D$, and we model them by $\M$ a submanifold of $\rrr^D$ to be estimated. Then $D$ is called the ambient
dimension (of the data). The ML algorithms that we will focus on can be seen as
finding an embedding $F:\M\rightarrow \rrr^m$, with $m\geq d$ and $m\ll D$;
in particular, if $m=d$, the embedding $F$ is a (global) coordinate chart. 

An advantage of embeddings is that one can avoid using multiple charts
to describe a manifold. Instead, one can find a global mapping
$F:\M\subset\rrr^D\rightarrow \N\subset \rrr^m$, where $\N$ is easier to understand. Whitney's
embedding Theorem \citep{smoothmfd} states that every $d-$dimensional manifold can be
embedded into $\rrr^{2d}$. Therefore, if one can find a valid
embedding, a significant dimension reduction can be achieved (from $D$
to $O(d)$). This is one of the major targets of manifold learning
algorithms.

This section has introduced manifolds as spaces that are ``like
$\rrr^d$'' locally around a point $\pb$. Next, we show how concepts
such as distances and angles, that is, Euclidean geometry, are
transferred from $\rrr^d$ to $d$-manifolds.

\subsection{(Riemannian) geometry on manifolds and isometric embedding}
\label{sec:def-rmetric}
\begin{figure}
\centering
\begin{subfigure}[b]{0.45\textwidth}
\includegraphics[width=\textwidth]{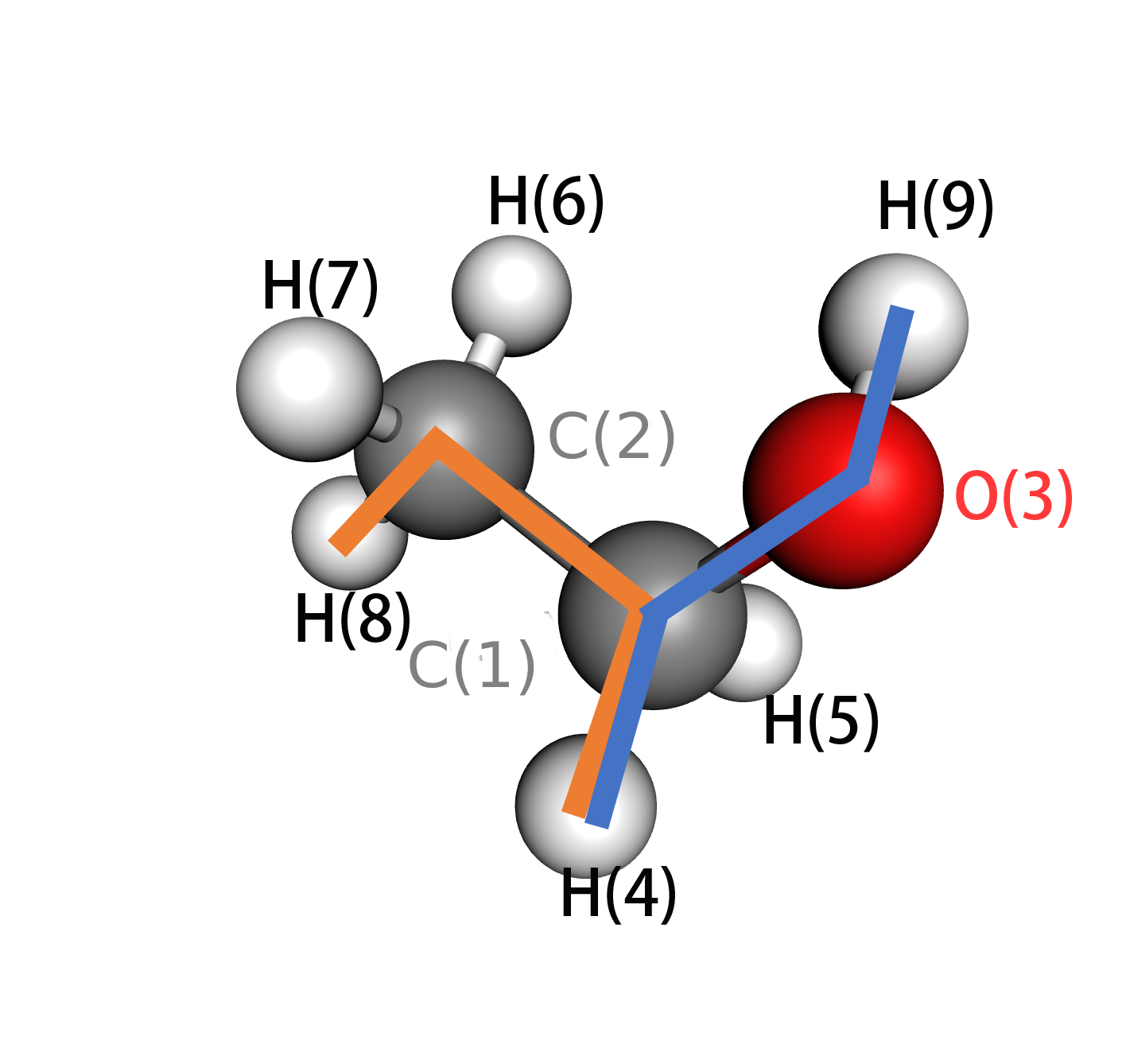}
\end{subfigure}
\begin{subfigure}[b]{0.45\textwidth}
\includegraphics[width=\textwidth]{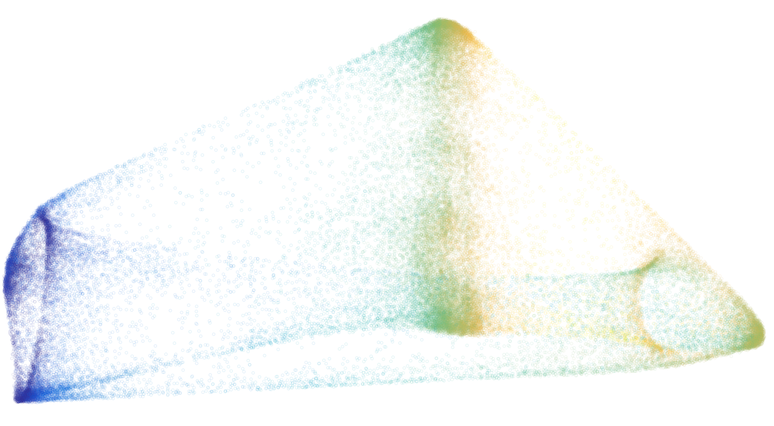}
\end{subfigure}
\caption{\label{fig:ethanol}{\bf Left:} The ethanol molecule has 9 atoms; a spatial configuration of ethanol has $D=3\times 9$ dimensions. The CH$_3$ group (atoms 2,6,7,8) and the OH group (atoms 3,9) can rotate w.r.t. the middle group (atoms 1,4,5), and the blue and orange lines represent these angles of rotation. {\bf Right:} A 2-manifold estimated from 50,000 configurations of the ethanol molecule. The manifold has the topology of a torus, and the color represents the rotation of the OH group. The sharp ``corners'' are distortions introduced by the embedding algorithm (explained in Section \ref{sec:graph}). Figure \ref{fig:graph-density-effects} shows the original data. This dataset is from \citet{chmielaTkaSauceSchuPMull:force-fields17}}

\end{figure}

For the data in Figure \ref{fig:ethanol}, a scientist may be
interested in the distance between two molecular configurations
$\xb_1,\xb_2$, seen as points of $\M\subset \rrr^D$. Their Euclidean
distance $\|\xb_1-\xb_2\|$ is readily available without requiring any
additional statistics. However, this value may not be of physical
interest, since most of the putative configurations along the segment
$\xb_1$ to $\xb_2$ in $\rrr^D$ are not physically possible. To deform
from state $\xb_1$ to $\xb_2$, the ethanol molecule must follow a path
contained in (or near) the manifold $\M$ of possible configurations,
and the distance $d_\M(\xb_1,\xb_2)$ shall naturally be defined as the
shortest possible length of such a path (and is called the
\mydef{geodesic distance}). Just like in $\rrr^d$ the distance
between two points is independent of the choice of basis, and
invariant if $\rrr^d$ is a subspace of a larger Euclidean space,
distances along curves in a manifold $\M$ can be defined solely based
on the coordinate charts $(U,\phi)$, hence \myemph{intrinsically},
without reference to the ambient space $\rrr^D$, and are purely
geometric, hence are independent of the choices of charts. This is
achieved through Riemannian geometry, as follows.

In $\rrr^d$, the scalar product $\langle \vb,\ub\rangle=\vb^T\ub$ is
sufficient to define both distances, by $\|\vb-\ub\|^2=\langle
\vb-\ub,\vb-\ub\rangle$, and angles, by $\angle (\vb, \ub)=
\cos^{-1}(\langle \vb,\ub\rangle/(\|\vb\|\|\ub\|))$. Moreover, any
positive definite matrix $\Ab\in \rrr^{d\times d}$ can induce an
inner product by $\langle \vb,\ub\rangle_{\Ab}=\vb^T\Ab\ub$; in
this context, $\Ab$ is often called a \mydef{metric} on $\rrr^d$.

Riemann took up this idea and introduced the \mydef{Riemannian
  metric}, which plays the same role as $\Ab$ above; however, this
metric is allowed to vary from point to point, smoothly.
\mmp{NOT NEEDED First, one needs to find spaces ``like $\rrr^d$'' which are associated
to each point $\ponM$. Then, one defines scalar products on them, via
positive definite matrices $\Ab$.}

\paragraph*{Tangent space and Riemannian metric}
The {\em tangent space} $\tp$ at a point $\pb \in \M$ is a
$d$-dimensional vector space of {\em tangent vectors} to $\M$. The
canonical basis of $\tp$ is given by the tangents to the coordinate
functions seen as curves on $\M$, while the tangent vectors can be seen
as tangents (or velocity vectors) at $\pb$ to smooth curves on $\M$
passing through $\pb$.  The collection of tangent spaces $\tp$ for all
points $\pb\in\M$, is called the tangent bundle of $\M$, denoted by
$\T\M$.  

A {\em Riemannian metric} $\g$ of a manifold $\M$ associates to each point
$\ponM \in \M$ an inner product $\langle \cdot,\cdot
\rangle_{\g(\ponM)}$ on the tangent space $\tp$, which varies smoothly
on $\M$.
%
The inner product $\g$ defines on each tangent space the norm $\lVert
\tgvec\lVert_{\g} = \sqrt{\langle \tgvec,\tgvec\rangle_{\g}}$,
distance $\norm{\tgvec_1-\tgvec_2}_{\g}$, angle $\cos^{-1}(\langle
\tgvec_1,\tgvec_2\rangle_{\g}/(\norm{\tgvec_1}_{\g}\norm{\tgvec_2}_{\g}))$
for all vectors in $\tp$. More importantly, infinitesimal quantities
such as the line element
$\mathrm{d}l=\sum_{i,j=1}^d\g_{ij}\mathrm{d}x^i\mathrm{d}x^j$ and
volume element
$\mathrm{d}V=\sqrt{\det(\g)}\mathrm{d}x^1\cdots\mathrm{d}x^d$ are also
expressed through the Riemannian metric, allowing one to define
lengths of curves and volumes of subsets of $\M$ as integrals. These
integrals are invariant to the choice of bases in $\T\M$, hence to the
choice of coordinate charts on $\M$. \mmp{say something about uniform
  distribution on manifold with finite volume?}

\mmp{In the special case $\M=\rrr^d$, with $\tp=\rrr^d$ for all
$\pb\in\rrr^d$ and the canonical basis, the Riemannian metric at every
point is the unit matrix, i.e., $\g_{\pb}\equiv \bm{I_d}$.}

\paragraph*{Isometry and isometric embedding}
A smooth map $F:\M\rightarrow \N$  induces  linear maps $\dddF
_{\pb}:\tp\rightarrow \T_{F(\pb)}\N$ called the {\em differential} of
$F$ at $\pb$. If we fix the coordinate systems on $\M$ and $\N$,
$\dddF_{\pb}$ becomes a $\dim \N\times \dim \M$ matrix which maps 
$\vb\in\tp$ to $\dddF_{\pb} \vb \in \T_{F(\pb)}\N$ (i.e., the {\em Jacobian} of $F$ in the given coordinates).

A smooth map $F:\M\rightarrow\N$ between Riemannian manifolds
$(\M,\g),(\N,\mathbf{h})$, is an {\em isometry} if the Riemannian
metric $\g$ at each point $\ponM$ is preserved by $F$, i.e. iff
\begin{equation}    \label{eq:defi-isometry}
  \text{for all \ }\pb\in \M \text{ and }
    \tgvec_1,\tgvec_2 \in \tp,\quad
    \langle \tgvec_1,\tgvec_2 \rangle_{\g(\ponM)} = \langle
    \dddF_{\ponM}(\tgvec_1),\dddF_{\ponM}(\tgvec_2)
    \rangle_{\mathbf{h}(F(\ponM))} 
\end{equation}
An isometry $F$ preserves geometry quantities such as angles,
distances, path lengths, volumes etc. \mmp{If \ref{eq:defi-isometry}
  does not globally hold, but holds in a neighborhood for each
  $\tp\in\M$. -- needed? reformulate} An embedding that is also an
isometry is called an {\em isometric embedding}.

\mmp{For manifold learning, we almost always want to assume that the
manifold where data lie is locally isometrically embedded in the
ambient space $\rrr^D$. Since as we will see, these algorithms use
local geometry in $\rrr^D$ to approximate the intrinsic geometry.}

Ideally, we would like a manifold learning algorithm to
produce an embedding $F$ into $\rrr^m$ that is isometric. Here we face
one of the most remarkable gaps between mathematical theory and
methodology in manifold learning. Although it was long proved
(Nash embedding theorem \citep{smoothmfd}) that isometric embedding is possible, no known practical
algorithm capable of isometric embedding exists at this time (details
and refinement of this statement are in Section
\ref{sec:geo-isometric}). However, by estimating auxiliary
information, working with a non-isometric embedding as if it was
isometric is still possible (see Section \ref{sec:rmetric-alg}).

%% file: rev-ml.tex
\paragraph*{The Manifold Assumption} 
Suppose we are given data $\{\bm{x}_i\}_{i=1}^n$ where each data entry
$\bm{x}_i\in\rrr^D$. It is assumed that data are
sampled from a distribution $\mathbb{P}$ that is supported on, or close to a
$d$ dimensional manifold $\M$ embedded in $\rrr^D$. This is the {\em
Manifold Assumption}. Throughout this survey, with a few noted exceptions, we will discuss the no noise case, when the data lie exactly on $\M$. \mmp{However, empirically, ML algorithms are robust to noise and are excellent for smoothing high dimensional data.}

\paragraph*{Manifold learning} A manifold learning algorithm can be thought as a mapping $F$ of $\bm{x}_i\in\rrr^D$ to
$y_i\in\rrr^m$. The {\em embedding dimension} $m$ is usually much
smaller than $D$ but could be higher than the intrinsic dimension
$d$. In the regime that $\mathbb{P}$ is supported exactly on $\M$, and
sample size $n\rightarrow \infty$, a valid manifold learning algorithm
$F$ should converge to a smooth embedding function $F$. This implies
that the algorithm should be guaranteed to recover the manifold
$\M$ \mmp{?? as long as $\M$ is not singular (as will be defined
later)}, regardless of the shape of $\M$.

Once the manifold assumption is believed to be true, most manifold
learning and non-linear dimension reduction methods can be grouped
into three paradigms, which differ in the way they represent the recovered
manifold. They are local linear approximations
(Section \ref{sec:lpca}), Principal Curves and Surfaces
(\ref{sec:pcs}), and embedding algorithms, which will be the focus of
Section \ref{sec:algs}. 

\begin{table} 
\caption{\label{tab:ml-paradigms}Three main paradigms for non-linear dimension reduction}
\begin{tabular}{llllll} 
{\bf Paradigm} & {\bf Representation} & {\bf }\\
\hline
{\bf Linear local}
& $\xb_i \in
U\subset \M\;\stackrel{F}{\rightarrow}\;\vb_i\in\hat{\T}_{\pb}\M$
$\cong \rrr^d$
& $D\rightarrow d$, local coordinates only
&
\\
{\bf Principal Curves}
&$\xb_i \in \rrr^D\;\stackrel{F}{\rightarrow}\;\xb'_i\in\M\subset\rrr^D$
&$D\rightarrow D$, global coordinates, noise removal
&
\\
{\bf  and Surfaces}\\
{\bf Embedding}
& $\xb_i \in \subset \M\;\stackrel{F}{\rightarrow}\;\yb_i\in F(\M)\subset \rrr^m$
& $D\rightarrow m$, with $m\geq d$, global coordinates (or charts)
&
\\
\end{tabular}
\end{table}


The Manifold Assumption itself is testable. For example in \citet{Fefferman2016-bc}, tests whether, given an i.i.d. sample, there exists a manifold $\M$ that can approximate this sample with tolerance $\varepsilon$\mmp{$\int d(\M,x)^2 < \varepsilon$}. These results are currently not practically useful, as knowledge of usually unknown parameters of the manifold ($d$, reach, volume) must be known or estimated. However, they, as well as \cite{GenovesePVW12}, give us the confidence to develop and use ML algorithms in practice.\mmp{more to add -- elsewhere/not now: kitty's and hari's principal curves work, }

\subsection{Neighborhood graphs}
\label{subsec:nn}
Practically all manifold learning algorithms start with finding the
neighbors of each data point $\xb_i$. This leads to the construction of
a {\em neighborhood graph}; this graph, with suitable weights,
summarizing the local geometric and topological information in the
data, is the typical input to a non-linear dimension reduction
algorithm. Every data point $\xb_i$ represents a node in this graph,
and two nodes are connected by an edge if their corresponding data points are neighbors. Throughout the survey, we use
$\neigh_i$ to denote the neighbors of $\bm{x}_i$ and $k_i=|\neigh_i|$ be
the number of neighbors of $\bm{x}_i$ (including $\bm{x}_i$ itself). The matrix
$\mathbf{N}_i\in\mathbb{R}^{k_i\times D}$ is the matrix with each row
representing a neighbor of $\bm{x}_i$.

There are two usual ways to define neighbors. In
a \mydef{radius-neighbor graph}, $\bm{x}_j$ is a neighbor of
$\bm{x}_i$ iff $\norm{\bm{x}_i-\bm{x}_j}\leq r$. Here $r$ is a parameter that
controls the neighborhood scale, similar to a bandwidth parameter in
kernel density estimation. Consistency of manifold learning algorithms
is usually established assuming an appropriately selected neighborhood
size, that decreases slowly with $n$ (see
Section \ref{sec:epsilon}).  In the \mydef{$k$-nearest neighbor ($k$-NN)
graph}, $\bm{x}_j$ is the neighbor of $\bm{x}_i$ iff $\bm{x}_j$ is among the
closest $k$ points to $\bm{x}_i$. Since this relation is not symmetric, usually the neighborhoods are symmetrized to obtain an undirected neighborhood graph. 

The $k-$NN graph has many computational
advantages w.r.t. the radius neighbor graph; it is more regular and often it is connected when the latter is not.  More software is
available to construct (approximate) $k$-NN graphs fast for large
data. But theoretically, it is much more difficult to analyze, and
fewer consistency results are known for $k-$NN graphs (Sections \ref{sec:graph}, \ref{sec:laplacian}).
Intuitively, $k_i$ the number of neighbors in the radius graph is proportional to the local data density, and manifold estimation can be analyzed through the
prism of kernel regression; while the $k$-NN graph
is either asymmetric, or if symmetrized, becomes more complicated to
analyze. 

The distances between neighbors are stored in the distance matrix $\mathbf{A}$, with $\mathbf{A}_{ij}$ being the
distance $\norm{\bm{x}_i-\bm{x}_j}$ if $\bm{x}_j\in \neigh_i$, and  infinity
if $\bm{x}_j$ is not a neighbor of $\bm{x}_i$.

Some algorithms weight the neighborhood graph by weights that are non-increasing with distances; the resulting $n\times n$ matrix is called
the \mydef{similarity matrix} (or sometimes {\em kernel matrix}). The
weights are given by a \mydef{kernel function}, 
\begin{equation}    \label{eq:similaritymat}
    \mathbf{K}_{ij}:=\begin{cases}
        K\left(\frac{\norm{\bm{x}_i-\bm{x}_j}}{h}\right),\quad &\bm{x}_j\in\neigh_i,\\
        0,\quad &\text{otherwise.}
    \end{cases}
\end{equation}

The kernel function here is
almost universally the Gaussian kernel, defined as $K(u)=\exp(-u^2)$ (\citet{belkin-nyogi,TingHJ:10,coifman:06,Singer2011VectorDM},etc) .
In the above, $h$, the kernel width, is another hyperparameter that
must be tuned. Note that, even if $\neigh_i$ would trivially contain
all the data, the similarity $\Kb_{ij}$  vanishes for
far-away data points. Therefore, equation \eqref{eq:similaritymat} effectively
defines a radius-neighbor graph with $r\propto h$. Hence, a rule of
thumb is to select $r$ to be  a small multiple of $h$ (e.g.,
$3$--$10h$). \mmp{Something about kernels, with citations?}

It is sometimes also useful to have kernel function $K(u)=\mathbf{1}$. Then the similarity matrix $\Kb$ is the same as the unweighted adjacency matrix of the radius neighbor graph.  By construction, $\mathbf{K}$ is usually a
sparse matrix, which is useful to accelerate the computation.

\mmp{computationally, sparse important; theoretically, all kernels the same to expand}
\mmp{denote the two K's differently with superscripts?}

If the data is so that $D$ is large, which is often the case, and $n$
is large too, which is necessary for manifold recovery,
constructing the neighborhood graph can be the most
computationally demanding step of the algorithm. Fortunately, much work has
been devoted to speeding up this task, and approximate algorithms are
now available, which can run in almost linear time in $n$ (computation complexity can be reduced to $O(n^{1+\delta})$, where $\delta < 1$ is a positive constant)  and have very good accuracy (\cite{ram2010ltaps}).  

Next, we briefly describe two of the paradigms in
Table \ref{tab:ml-paradigms}, then, from Section \ref{sec:algs} on
focus on the third, embedding algorithms.

\subsection{Linear local approximation}
\label{sec:lpca}

This idea stems from the classical Principal Component Analysis, and it seeks to adapt it to data sampled from a curved
manifold, instead of a linear subspace.  Random projections \citep{Baraniuk2009,Chinmay2007} were first 
proposed to learn the structure of a low dimensional manifold. Let
$F:\rrr^D\rightarrow \rrr^m$ be a random orthogonal
projector\footnote{Obtained by orthogonalizing a matrix with i.i.d. normal or
Bernoulli entries}, it is sufficient to have $m\geq O(d\log
n/\varepsilon^2)$ to preserve all distances approximately, in the sense that
$(1-\varepsilon)\sqrt{m/n} \leq \dist(F(\xb_i),F(\xb_j))/ \dist(\xb_i,\xb_j)\leq
(1+\epsilon)\sqrt{m/n}$ holds for any $\xb_i,\xb_j$ from $\M$. Here
$\dist$ can be either geodesic distance on $\M$ or Euclidean distance
on $\rrr^D$. For large data, however, this approach leads to a large $m$, not useful for dimension reduction. 

Both PCA and random projections seek global linear representations, and they do not utilize the geometric structure of the manifold in a finer scale at a reference point $\bm{x}$. One improvement is to perform PCA on a weighted covariance matrix, with weights decaying away from $\bm{x}$,i.e. let

\beq 
\mathbf{C}=\frac{1}{n}\sum_{i=1}^n w_i(\xb_i-\bm{x})(\xb_i-\bm{x})^\top\,.
\eeq 
The weights can be computed using the kernel $K()$ and bandwidth $\bw$ used for the neighborhood graph, for example. This weighted PCA procedure is sometimes termed {\em local PCA (lPCA)}. Local PCA is only used to understand geometric structure near any reference point, usually one of the points in $\dataset$. One can map all the data by performing lPCA with a subset of the data points so that each $\xb_i$ is sufficiently well approximated by its projection. This way, all the data are represented in $d$ dimensional coordinates. However, some points can have more than one representation, if they are close to 2 or more reference points. All the representations (i.e., coordinates) are {\em local}, and understanding or making inferences on the entire manifold are tedious at best. \mmp{there is a paper to cite here, but i couldn't find it now}. This approach can be refined into a multiscale local linear approach, which is more accurate and parsimonious \citep{Chen2013}.

\subsection{Principal curves and principal $d$-manifolds}
\label{sec:pcs}
This paradigm is the only one of the three where noise is assumed. Consider data of the form $\xb_i=\xb^*_i+\epsilon_i$, where $\epsilon_i$ represents 0-mean noise, and the $\xb^*_i$ are sampled from a curve, for instance. Then, to recover the curve, one would need to estimate the local mean of the data, and this was proposed in the seminal work of \citet{hastiewerner1989}; unfortunately, the estimation is difficult, and this definition does not lead to a unique curve.
A more recent proposal that removed the previous difficulties was to define the principal curve as the {\em ridge} of the data density.
\begin{marginnote}A point $\xb$ is on a ridge if it is a local maximum of the density $f_{\xb}$ in $D-1$ directions, and the remaining direction coincides with the gradient $\nabla f_{\xb}$.
\end{marginnote}
Several ridges may meet at a peak, i.e., at a local maximum, but in between, the ridges are manifolds \citep{yen-chi} if the density $f_{\xb}$ is smooth enough. This concept can be extended to principal surfaces, and principal $d$-manifolds.

The ridge can be estimated with the {\em Subspace Constrained Mean Shift (SCMS)} algorithm (for more details, we refer the reader to \cite{OzertemErdogmus11}). The SCMS algorithm maps each $\xb_i$ to a point $\yb_i\in \rrr^D$ lying on the principal curve (or $d$-manifold). Hence, as a manifold estimation algorithm, this method does not reduce dimension. However, unlike local linear maps, the output is in a global coordinate system, i.e. $\rrr^D$.

Usually, the ridge does not coincide with the mean of the data; the bias depends on the curvature of the manifold: the density is higher on the ``inside'' of the curve. However, for their smoothing property, principal $d$-manifolds are remarkably useful in the analysis of manifold estimation in noise \citep{GenovesePVW12,mohammedNarayanan:localpcs17}.

\mmp{Remove:
Principal curves noise is implicitly assumed. Easy to map a new $\bm{x}$ to
the estimated $\M$ (interpolation). Disadvantages:
$F:\rrr^D\rightarrow \rrr^D$, hence dimension reduction is only
implicit. The algorithm is computationally expensive but can be
accelerated \cite{alon-thesis}. Typically used for $d=1$; was recently
extended \cite{yen-chi} to estimating submanifolds on the $d$-sphere;
hence to work with \mydef{directional data}.}

We have quickly reviewed two simple methods for manifold estimation: local linear approximation reduces the dimension locally, but offers no global representation, while principal curves produce a global representation but do not reduce dimension. It is time to focus on the third class, of algorithms that produce embeddings, representations that are both global and low dimension.

%% file: rev-algs.tex
The term "manifold learning" was proposed in the seminal work of two algorithms: LLE(\cite{roweis:00}) and {\sc Isomap}(\cite{tenenbaum00:isomap}), together inaugurated
the modern era of non-linear dimension reduction. In this subsection,
we introduce classical manifold learning algorithms that (attempt to)
find a global embedding $\Yb\in \rrr^{n\times m}$ of data set $\dataset$. 

We separate the algorithms, roughly, into ``one-shot'' algorithms,
which obtain embedding coordinates from the principal eigenvectors of some
matrix derived from the neighborhood graph, or by solving some other
global (usually convex) optimization problem,  and ``attraction-repulsion''
algorithms, which proceed from an initial embedding $\Yb$ (often
produced by a one-shot algorithm) and improve
it iteratively. While this taxonomy can rightly be called superficial,
at present, it represents a succinct and relatively accurate summary of
the state of the art.

No matter what the approach, given the neighborhood information
summarized in the weighted neighborhood graph, an embedding
algorithm's task is to produce a smooth mapping of the inputs which
distorts the neighborhood information as little as possible. The
algorithms that follow differ in their choice of information to
preserve, and in the sometimes implicit constraints on smoothness. 

\subsection{Review: Principal Component Analysis (PCA)}
Before we dive deep into various non-linear dimension reduction algorithms, we have a brief review on linear dimension reduction methods.

Linear dimension reduction methods find global embedding of the data in a low dimensional linear subspace. One way of understanding principal component analysis is to find $d$ dimensional linear subspace $\mathcal{V}$ such that the data $\{\xb_i\}$ projected onto it have the smallest reconstruction error. If $\mathcal{V}$ has an orthogonal basis $\mathbf{T}\in\rrr^{D\times d}$ such that $\mathbf{T}^\top\mathbf{T}=\mathbf{I}_d$. Then $\xb_i$ projected onto $\mathcal{U}$ has low dimensional representation $\yb_i=\mathbf{T}^\top\xb_i$ under basis $\mathbf{T}$. In $\rrr^D$, projection of $\xb_i$ onto $\mathcal{U}$ is given by $\mathbf{T}\mathbf{T}^\top \xb_i$. If we introduce the data matrix $\mathbf{X}\in\mathbf{R}^{n\times D}$, with $i-$th row being $\xb_i^\top$, then the low dimensional representation matrix $\Yb\in\rrr^{n\times d}$ is given by $\mathbf{X}\mathbf{T}$ and in $\rrr^D$ the projected data matrix is $\mathbf{X}\mathbf{T}\mathbf{T}^\top$.

Then we can write PCA problem as
\begin{equation} \label{eq:pca}
    \min_{\mathbf{T}:\mathbf{T}\in\rrr^{D\times d},\mathbf{T}^\top\mathbf{T}=\mathbf{I}_d} \sum_{i=1}^n\norm{\xb_i-\mathbf{T}\mathbf{T}^\top\xb_i}^2= \min_{\mathbf{T}:\mathbf{T}\in\rrr^{D\times d},\mathbf{T}^\top\mathbf{T}=\mathbf{I}_d}\norm{\mathbf{X}-\mathbf{X}\mathbf{T}\mathbf{T}^\top}_F^2
\end{equation}

 Consider the singular value decomposition preserving only the first $d$ singular values of $\mathbf{X}=\mathbf{U}\bm{\Sigma}\mathbf{V}^\top$ where $\mathbf{U}\in\mathbb{R}^{n\times d},\mathbf{V}\in\mathbb{R}^{D\times d}$ are orthogonal matrices and $\bm{\Sigma}$ is $d\times d$ diagonal matrix, then solution of this problem is $\mathbf{T}=\mathbf{V}$. Low dimensional representation of original data is $\mathbf{Y}=\Xb\mathbf{T}=\mathbf{U}\bm{\Sigma}$, these are also called principal components. In the terminology of PCA, columns of $\mathbf{V}$ are called principal vectors. They characterize the directions that explain the variance in the data. 

When data $\xb_i$ are centered, the (unnormalized) sample covariance matrix of data is $\mathbf{C}=\mathbf{X}^\top\mathbf{X}$. The solution to PCA can also be found by eigendecomposition of $\mathbf{C}$. The first $d$ eigenvectors of $\mathbf{C}$ is just matrix $\mathbf{V}$. If the dimension $D\gg n$, it will be easier to first compute the Gram matrix $\mathbf{C}=\mathbf{X}^\top\mathbf{X}$ and then perform a truncated eigen-decomposition $\mathbf{C}=\mathbf{V}\bm{\Sigma}^2\mathbf{V}^\top$. Low dimensional representation is still $\Xb\mathbf{V}$.

\subsection{``One shot'' embedding algorithms}
\label{sec:emb-algs}

\subsubsection{Isomap}
\label{sec:isomap}
Classical Multidimensional Scaling (MDS,\citet{MDS}) takes input from a pairwise distance matrix $\Ab$ and outputs coordinates $\Yb\in\rrr^d$ that best preserve the distances. {\sc Isomap} is a generalization of multidimensional scaling that preserves distances between data points while finding low dimensional coordinates. Instead of Euclidean distance in classical MDS, {\sc Isomap} use shortest path distances in the neighborhood distance graph to approximate geodesic distance on a manifold.

\begin{algorithm}
\caption{{\sc Isomap}}
\label{alg:isomap}
\begin{algorithmic}[1]
\REQUIRE: Neighborhood distance matrix $\mathbf{A}$, embedding dimension $m$ 
\STATE Compute shortest path distance matrix $\tilde{\Ab}_{ij}$::
\begin{equation*}
    \tilde{\Ab}_{ij}=\begin{cases}
    \Ab_{ij} &\quad \mathbf{A}_{ij}<\infty\;,\\
    \text{shortest path distance between $i,j$} &\quad \mathbf{A}_{ij}=\infty\;.
    \end{cases}
\end{equation*}
\STATE Multidimensional Scaling  $\Yb=\operatorname{MDS}(\mathbf{M},d)$ with $\bw{M}=[\tilde{\Ab}_{ij}^2]$
\ENSURE: $m$ dimensional coordinates $\mathbf{Y}$ for $\dataset$
\end{algorithmic}
\end{algorithm}
Intuitively, the shortest
graph distance is a good approximation to the geodesic distance in a
neighborhood provided that data are sufficiently dense in this region
and neighborhood size is appropriately chosen~\citep{bernsteinDeSilvaLangfordTenn:00}. \hanyuz{Find a precise
  statement here, the original paper statement is ambiguous and
  geometric assumptions are not clear.} In the limit of large $n$, {\sc Isomap} was shown to produce isometric embeddings for $m=d$, whenever the data manifold is {\em flat}, i.e. admits an isometric embedding \mmp{is this so??} in $\rrr^d$, and data space is convex. Empirically, {\sc Isomap} embeddings are close to isometric also when $m>d$ and $m$ is sufficient for isometric embedding. 

Computation complexity of {\sc Isomap} is $O(n^3)$, with a most computational burden for computing all pairs of shortest path distance. Space complexity is $O(n^2)$. Since {\sc Isomap} works with dense matrices, this space complexity cannot be improved.

\mmp{Landmark isomap -- i don't have citation, to make it fast}

There are variants of Isomap that improve it in different ways: Hessian Eigenmap \citep{Donoho2003HE} enables non-convex data where they introduce the use of Hessian operator; Continuum Isomap \citep{Zha2007ContIsomap} generalizes Isomap to a continuous version such that out-of-sample extension of Isomap is possible.\mmp{and Landmark Isomap~\cite{} accelerates it by only computing a subset of the shortest paths, and approximating the rest. -- if we find it}

\subsubsection{{Diffusion Maps}/{Laplacian Eigenmaps}}
\label{sec:diffmap}
Unlike {\sc Isomap}, \dmalg, as well as most embedding methods, work with a sparse matrix derived from the similarity $\Kb$; namely, they embed the data by the eigenvectors of the {\em graph Laplacian}  $\Lbw$. The construction of $\Lbw$, also called the {\em {Diffusion Maps} Laplacian} or {\em renormalized Laplacian}, is described in Algorithm \ref{alg:renormLap} (and it consists of a column normalization of $\Kb$, followed by a row normalization). The \LEalg~algorithm differs from \dmalg~only in the use of a different Laplacian, $\Lbw^{norm}$ below. 

\begin{algorithm}
\caption{{\sc Diffusion Maps}/{\sc Laplacian Eigenmaps}}
\label{alg:laplacianeigenmap}
\begin{algorithmic}[1]
\REQUIRE Graph Laplacian $\Lbw$ (or $\Lbw^{nor}$, embedding dimension $m$.
\STATE Compute eigenvectors of smallest $m+1$ eigenvalues of $\Lbw$, $\{\bm{v}^{i}\}_{i=0}^{m}$, each eigenvector $\bm{v}^i\in\mathbb{R}^n$.
\STATE Discard $\bm{v}^0$.
\STATE Represent each $\bm{x}_j$ by $\bm{y}_j=(v^1_j,\cdots,v^m_j)^\top$ \label{step:evecs}
\ENSURE $\mathbf{Y}$
\end{algorithmic}
\end{algorithm}
To construct a graph Laplacian matrix, let
$d_i=\sum_{j\in\neigh_i}\mathbf{K}_{ij}$ represent the degree of node
$i$ and $\mathbf{D}=\diag\{d_1,\cdots,d_n\}$. Then multiple choices of
graph Laplacian exist:
\begin{itemize}
    \item Unnormalized Laplacian: $\Lbw^{un}=\mathbf{D}-\mathbf{K}$
    \item Normalized Laplacian: $\Lbw^{nor}=\mathbf{I}-\mathbf{D}^{-1/2}\mathbf{K}\mathbf{D}^{-1/2}$
    \item Random-walk Laplacian: $\Lbw^{rw}=\mathbf{I}-\mathbf{D}^{-1}\mathbf{K}$
    \item Renormalized Laplacian $\Lbw$ described below.
\end{itemize}

\begin{algorithm}
\caption{Renormalized Laplacian}
\label{alg:renormLap}
\begin{algorithmic}
\REQUIRE Neighborhood similarity matrix $\mathbf{K}$, kernel function $k()$ and kernel bandwidth $h$
\STATE Compute similarity matrix $\mathbf{K}_{ij}=k(\mathbf{A}_{ij}/h)$
\STATE Normalize columns: $d_j=\sum_{i=1}^n\mathbf{K}_{ij}$, $\;\tilde{\mathbf{K}}_{ij}=\mathbf{K}_{ij}/d_j$  for all $i,j=1,\ldots n$
\STATE Normalize rows: $d_i'=\sum_{j=1}^n\tilde{\mathbf{K}}_{ij}$, $\;\mathbf{P}_{ij}=\tilde{\mathbf{K}}_{ij}/d_i'$ for all $i,j=1,\ldots n$
\ENSURE  $\Lbw=(\mathbf{I}-\mathbf{P})/h^2$ 
\end{algorithmic}
\end{algorithm}
Why one Laplacian rather than another? The reason is that, even though in
many simple examples the difference is hard to spot, one needs to
ensure that, as more sample sizes are collected, the limit of these
$\Lbw$'s is well defined, and the embedding algorithm is
unbiased. \mmp{NOT TRUE?? Trillos. The lack of normalization makes
$\Lbw^{unnorm}$ diverge} It is easy to see that $\Lbw^{norm}$ and
$\Lbw^{rw}$ are similar matrices. Moreover, whenever the degrees $d_i$
are constant, $\Lbw=\Lbw^{rw}\propto \Lbw^{un}$, hence all Laplacians
should produce the same embedding. The difference appears when the
data density is non-uniform, making one $\xb_i$ be surrounded more
densely by other data points. The seminal work \citet{coifman:06}, which
introduced renormalization, showed that in this case, the eigenvectors
of $\Lbw^{norm}, \Lbw^{rw}$ are biased by the sampling density and
that renormalization removes this bias. Sections \ref{sec:laplacian}, and Figure \ref{fig:graph-density-effects} illustrate this. 


The idea of {\em spectral embedding} appeared in \mmp{earlier, unnormalized, for graph embedding}\citep{ShiMalik00,belkin:01} as a trick for clustering, and was then generalized
as a data representation method in \citet{belkin:03} as \LEalg. They
connect the graph Laplacian with the famous Laplace-Beltrami operator
$\Delta_{\M}$ of manifold, a differential operator that plays an
an important role in modern differential
geometry \citep{rosenberg_1997}. Estimating the Laplace-Beltrami operator itself is
an important geometric estimation problem that will be reviewed in
Section \ref{sec:laplacian}.

Because $\Lbw$ is sparse, \dmalg/\LEalg~ are computationally less challenging when compared with {\sc Isomap}.

\subsubsection{Local Tangent Space Alignment (LTSA)}
This algorithm, proposed in \citep{ZhangZ:04} seeks to find local representation in the tangent space at each point $\xb_i$, then aligns these to obtain global coordinates. \mmp{It assumes that the observed dataset $\dataset\{\bm{x}_i\}_{i=1}^n$ can be written as the image of some smooth mapping $F$, i.e., $\bm{x}_i=F(\yb_i)$.}

 The first stage of LTSA finds the local representation of neighboring points $j\in\neigh_i$ via projections on the tangent $\T_{\xb_i}\M$; thus $\yb_{j}-\yb_i$ can locally be approximated by an affine transformation of orthogonal projections of $\bm{x}_{j}$ onto tangent space at $\bm{x}_i$ through Taylor expansion. The optimal affine transformation is obtained by minimizing the reconstruction error near each $\bm{x}_i$
 \begin{equation}
     \min_{\tilde{\bm{x}_i},\Theta,\mathbf{Q}} \sum_{j\in\neigh_i}\lVert \bm{x}_j - (\tilde{\bm{x}}_i+\mathbf{Q}\theta_j^{(i)})  \lVert^2\;,
 \end{equation}
Where $\bm{x},\mathbf{Q}$ \mmp{what are the dimensions of these?} are translation and rotation that parametrize this affine transformation. $\theta_j$ is a local coordinate of each neighbor $\bm{x}_j$ projected on this linear subspace.

In the second stage of LTSA, one obtains global embedding coordinates $\Yb$ while\mmp{ maintaining} $\theta_j$ that preserves local geometry information, through minimizing a global reconstruction error

\begin{equation}
     \min_{\{\bm{y}_i\}_{i=1}^n,\{\mathbf{P}_i\}_{i=1}^n} \sum_{i=1}^n\sum_{j\in\neigh_i} \lVert \bm{y}_j -\tilde{\bm{y}}_i -  \mathbf{P}_i\theta_j^{(i)}\lVert^2
\end{equation}

The optimization in both steps can be transformed into eigenvalue problems. Hence the algorithmic procedure of LTSA is displayed in algorithm \ref{alg:ltsa}

\begin{algorithm}
\caption{ {\sc Local tangent space alignment}}
\label{alg:ltsa}
\begin{algorithmic}
\REQUIRE Dataset $\dataset$, embedding dimension $m$.
\STATE $\mathbf{B} = 0$
\FOR{$i=1,2,\cdots,n$}
\STATE Find the $k$ nearest neighbors of $\xb_i$: $\bm{x}_j,j\in\neigh_i$.
\STATE Find local dataset $\bm{\Xi}_i=[\bm{x}_j-\tilde{\bm{x}}_i]_{j\in\neigh_i}$, where $\tilde{\bm{x}}_i$ is the average of all neighbors of $\bm{x}_i$.
\STATE Compute the $m$ largest eigenvectors $\tilde{\bm{v}}^1,\cdots,\tilde{\bm{v}}^d$ of $\bm{\Xi}^\top\bm{\Xi}$, set $\mathbf{G}_i=[\mathbf{1}/\sqrt{k},\tilde{\bm{v}}^1,\cdots,\tilde{\bm{v}}^d]$
\STATE $\mathbf{B}=\mathbf{B}+\mathbf{I}-\mathbf{G}_i\mathbf{G}_i^\top$.
\ENDFOR
\STATE Compute the 2 to $m+1$ smallest eigenvectors of $\mathbf{B}$, $\{\bm{v}^j\}_{j=1}^{m}$, each eigenvector $\bm{v}^j\in\rrr^n$.
\ENSURE $m$ dimensional embeddings $\bm{y}_i=(v_i^1,\cdots,v_i^m)$ 
\end{algorithmic}
\end{algorithm}

 We have seen three embedding algorithms so far in this section: {\sc Isomap,LE,LTSA}, which all come with a certain performance guarantee. On the other hand, locally linear embedding (LLE, \citet{roweis:00}) is a heuristic method that utilizes a very similar idea: estimate local representations first and then align them globally. However, vanilla LLE does not perform well empirically and lacks theoretical guarantees \citep{TingHJ:10,Ting2018}.

%% file: rev-ies.tex
\subsection{``Horseshoe'' effects, neighbor embedding algorithms,  and selecting independent eigenvectors}
\label{sec:ies}
\mmp{neighbor embedding or attraction-repulsion?}
Algorithms that use eigenvectors, such as DM, are among the most
promising and well-studied in ML (see Sections \ref{sec:graph},\ref{sec:epsilon},\ref{sec:laplacian}). Unfortunately, such algorithms fail when the data manifold has a large aspect ratio (such as a
long, thin strip, or a thin torus).  This problem has been called \myemph{the Repeated Eigendirection
  Problem (REP)} and has been demonstrated for {\sc LLE, LE, LTSA, HE} \citep{GolZakKusRit08}, and is pervasive in real data sets.

From a differential geometric standpoint, the REP is a drop in the rank of the embedding Jacobian, due to eigenvectors (or eigenfunctions, in the
limit) that are harmonics of previous ones, as shown in Figure
\ref{fig:ies}. For example, for a rectangular strip (and a finite sample), the
scatterplot of $(\bw{v}^1_i,\bw{v}^2)_{i=1,\ldots n}$ from step \ref{step:evecs} of the \dmalg~follows a parabola; hence, it is a 1-dimensional mapping, even though the rectangle is 2-dimensional. This fact is a relevant diagnosis for REP in practice:
when an embedding looks like a ``horseshoe'', this may not represent
a property of the data, but an artifact signalling that one of the
data dimensions is collapsed, or poorly reflected in the embedding~\citep{diaconisGoelHolmes:horseshoes08}.

\subsubsection{Relaxation-based neighbor embedding algorithms}
\label{sec:neighbor-embed}
The pervasiveness of the REP stimulated the development of algorithms
that balance attraction between neighbors in the original space, with
repulsion between neighbors in the embedding space
 \citep{tsne,mcinnes2018umap,forceatlas,carreira-perp:10,imVermaBranson:18}. Usually, the embedding
 coordinates $\Yb$ are optimized iteratively until equilibrium is reached.

 The \tsnealg~algorithm~\cite{tsne}, one variant of which~\citet{bohmBehrensKobak:tsne22} we briefly describe here, exemplifies this approach.

{\sc Stochastic neighbor embedding (SNE)}, proposed in \citet{HintonRoweis2002}, change neighborhood relationship from a hard 0-1 coding to conditional
probabilities. The algorithm computes two sets of conditional
probabilities: $p_{ij}$ which models the probability of $\bm{x}_i,\bm{x}_j$
being neighbors (and is the algorithm input), and $q_{ij}$ that models
the probability of output points $\bm{y}_i,\bm{y}_j$ being neighbors. In \citet{tsne},
the authors proposed to use a Student-t distribution to model these
conditional probabilities and, as \tsnealg, this algorithm became widely used.

In more detail, from $p_{ij}$ and $q_{ij}$, \tsnealg~constructs two
similarity matrices; $\Vb$ is the similarity between data points, calculated as $\Vb=(\Db^{-1}\Kb+\Kb\Db^{-1})/(2n)$, where $\Kb$
denotes the $k$-nearest neighbor similarity matrix. The matrix
$\Wb=[\,\Wb_{ij}\,]b_{i,j=1}^n$ represents similarities in the
embedding space; $\Wb_{ij}=\frac{1}{1+\Ab^{out}_{ij}}$ where
$\Ab_{ij}^{out}=\|\yb_i-\yb_j\|^2$ is the squared distance matrix in the embedding space, a dense matrix.

The \tsnealg~algorithm starts with arbitrary coordinates
$\Yb\in\rrr^{n\times m}$, and iteratively updates them by gradient
descent to minimize the following loss function, which is akin to a cross-entropy \citep{HintonRoweis2002,tsne}.

\beq \label{eq:tsne-cost}
\loss^{\tsnealg}\;=\;-\sum_{i,j}\frac{1}{n}\Vb_{ij}\ln \Wb_{ij}+\ln
\sum_{ij} \Wb_{ij}.
\eeq
In the above, $\sum_{ij} \Wb_{ij}=w_{tot}$
normalizes the entries of $\Wb$ to 1. Thus, the original aim of
\tsnealg~is to match the (normalized) data weights by the
(normalized) embedding weights around each point, which motivates the
name \emph{Stochastic Neighbor Embedding} (SNE, \citet{HintonRoweis2002}).\mmp{tsne citation = vander maaten}.

Uniform manifold approximation and projection (UMAP, \citet{mcinnes2018umap}) is
another popular heuristic method. On a high level, UMAP minimizes the
mismatches between topological representations of high-dimensional
data set $\{\bm{x}_i\}_{i=1}^n$ and its low-dimensional embeddings
$\bm{y}_i$. Theories of UMAP are still very limited. 

\tsnealg~ has the advantage of being sensitive to local structure and
to clusters in data~\citep{lindermanSteinerb:simods19,kobakLindermanSteinerbKlugerBerens:20} (but does not explicitly preserve
the global structure).  We note that this propensity for finding
clusters comes partly from the choice of neighborhood graph (Section
\ref{sec:graph}). However, this is not the whole
story. Recently, it has been shown that this property stems from the
gradient of the loss function $\loss^{\tsnealg}$, which has the form
\beq
\fracpartial{\loss^{\tsnealg}}{\yb_i}\;=\;-\sum_{ij}\Vb_{ij}\Wb_{ij}(\yb_i-\yb_j)+
\frac{n}{\rho}\sum_{ij}\frac{\Wb_{ij}}{w_{tot}}(\yb_i-\yb_j).
\eeq
In the
above, the first term is an attraction between graph neighbors, while the second
represents repulsive forces between the embedded points $\yb_{1:n}$
\citep{bohmBehrensKobak:tsne22,zhangGilbertSteinerb:theforce22}.  Note the additional important parameter $\rho$,
which controls the trade-off between attraction and repulsion ($\rho$
corresponds to a version of the cost with $w_{tot}^{1/\rho}$ in the
second term). In \citet{bohmBehrensKobak:tsne22} it is shown that varying $\rho$ from small
to large values $\rho$ decreases the cluster separation, and makes the
embedding more similar to the \LEalg~embedding. Moreover, quite
surprisingly, \citet{bohmBehrensKobak:tsne22} shows that by varying $\rho$, the
\tsnealg~can emulate a variety of other algorithms, most notably.

\umapalg~\citep{mcinnes2018umap} and \forcealg~\citep{forceatlas}.\mmp{something about
  bohm}. Other works that analyze the attraction-repulsion behavior of
\tsnealg~are \cite{zhangSteinerb:forceful21}. \mmp{move to epsilon?} One yet unsolved issue with \tsnealg~is the
choice of the number of neighbors $k$. Most applications use the
default $k=90$ \citep{Policar19}; this choice, as well as other behaviors
of this class of algorithms, are discussed in \cite{zhangGilbertSteinerb:theforce22}.

Note also that since the REP can be interpreted as extreme
distortion, the \rralg~(\citet{PerraultM:arxiv1406.0118} in Section \ref{sec:rmetric}) can
also, be used to improve the conditioning of an embedding in an
iterative manner. 

Finally, in {\em Minimum Variance Unfolding (MVU)}~ , proposed in \citet{weinbergerSaul:MVU06, arias-castroPelletier:MVU13},
repulsion is implemented via a {\em Semidefinite Program}, hence the
embedding $\Yb$ is obtained by solving a convex optimization. This algorithm can be seen both as a one-shot and as an attraction-repulsion algorithm;
\citet{diaconisGoelHolmes:horseshoes08} show that MVU is related to the fastest mixing Markov chain on the
neighborhood graph. \mmp{why put it here and not in the first section? it seems that the algorithms here are more focused on pairwise distances in embedding than the other category. is it fair to say this?}
\newlength{\tttwi}
\setlength{\tttwi}{\textwidth}
\newlength{\mypicwi}

\begin{figure}
  \setlength{\mypicwi}{0.25\tttwi}
  \centering
    \begin{subfigure}[b]{\mypicwi}
      \centering
      \rule{\mypicwi}{0em}
      \rule{0em}{\mypicwi}
    \end{subfigure}
    \begin{subfigure}[b]{\mypicwi}
      \centering
      \includegraphics[width=\mypicwi]{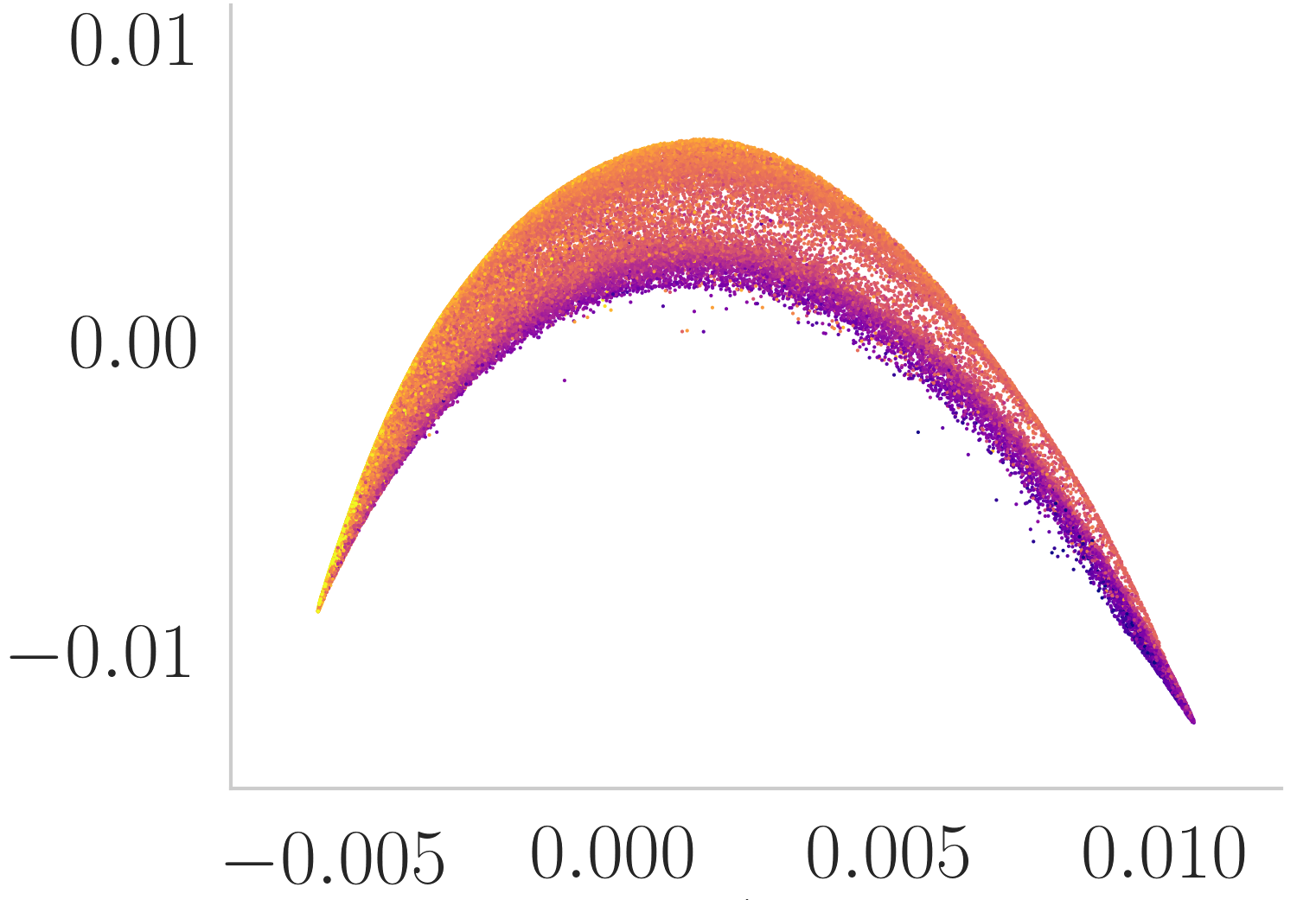} 
    \end{subfigure}
    \begin{subfigure}[b]{\mypicwi}
      \centering
      \includegraphics[width=\mypicwi]{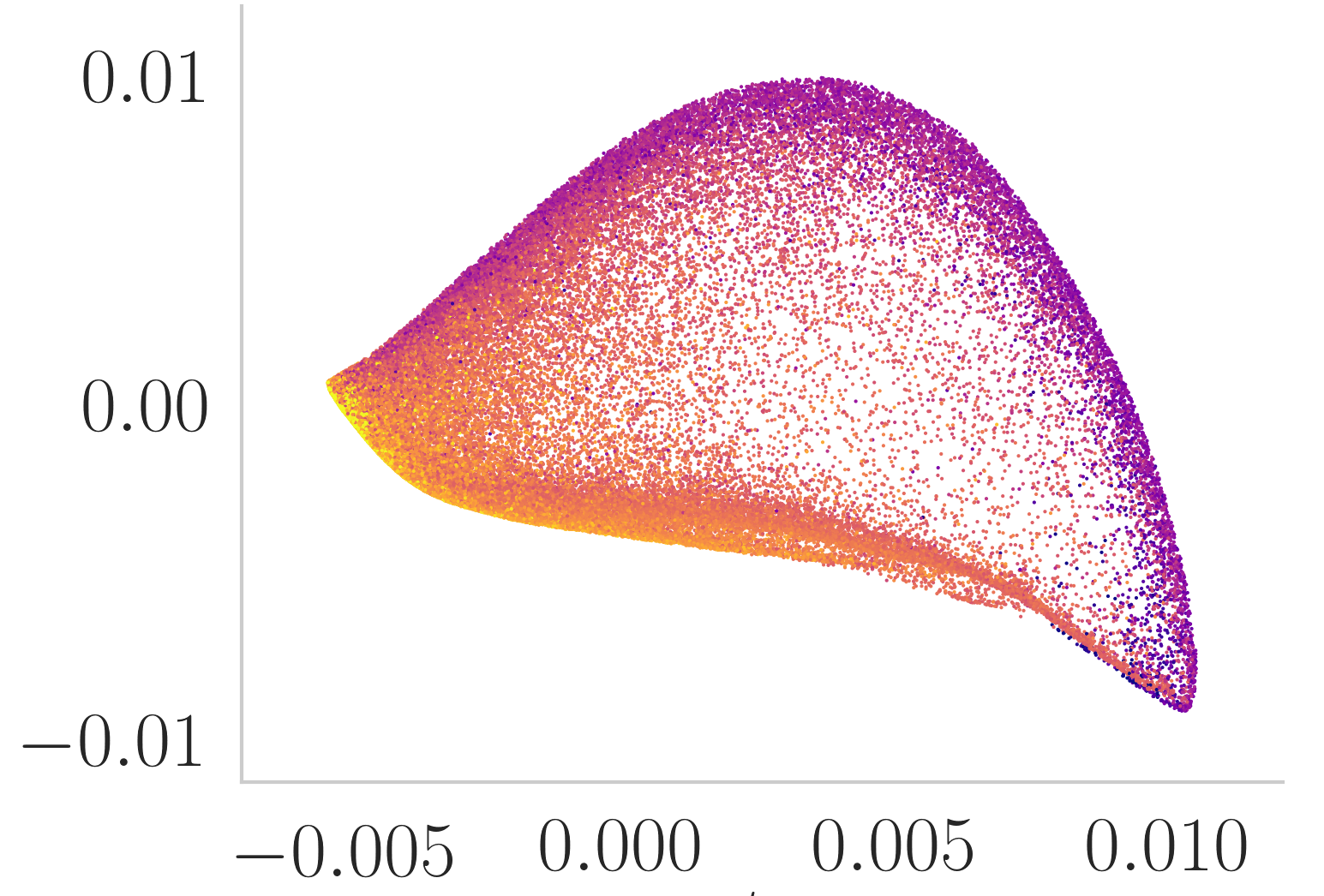}
    \end{subfigure}
    \\
    \begin{subfigure}[b]{\mypicwi}
       \centering
       \includegraphics[width=\mypicwi]{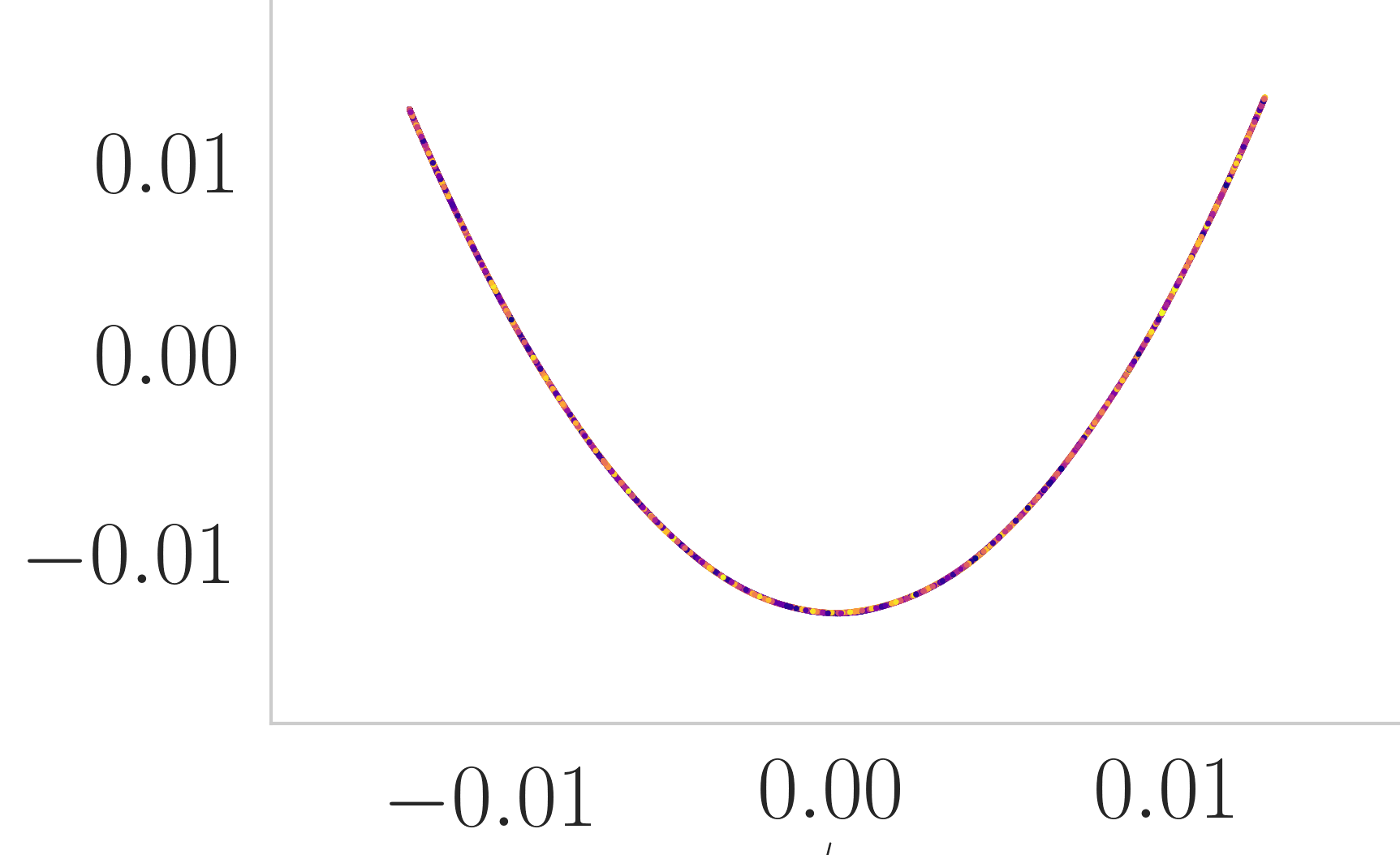}
    \end{subfigure}
    \begin{subfigure}[b]{\mypicwi}
       \centering
       \includegraphics[width=\mypicwi]{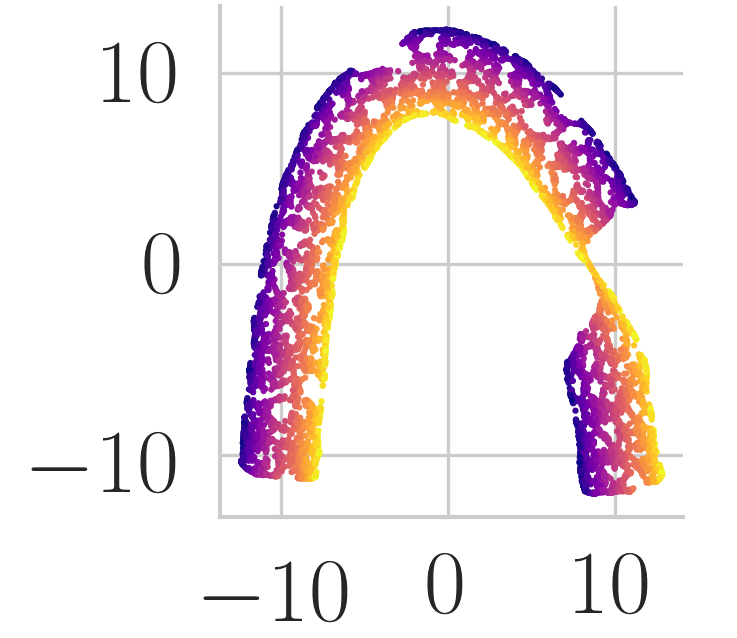}
    \end{subfigure}
    \begin{subfigure}[b]{\mypicwi}
       \centering
       \includegraphics[width=\mypicwi]{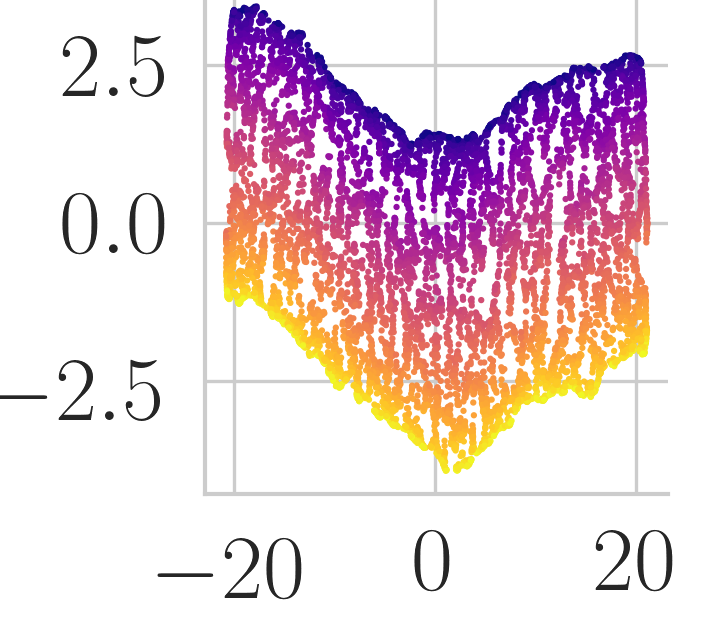}
    \end{subfigure}
    \caption{\label{fig:ies} Embedding algorithms failing 
      to find a full rank mapping, if they greedily select the first
      $m=2$ eigenvectors, and correction by a more refined choice of eigenvectors.  {\bf Top row}: Embeddings of galaxy spectra from the SDSS (Section \ref{sec:appli}) by \dmalg~; {\bf middle} ``horseshoe'' when first 2 eigenvectors are used; {\bf right} the same data, with selection of the
      second eigenvector (in this case by \cite{yuchaz}). {\bf Bottom row}: embeddings of a swiss roll with length 7 times the width. {\bf Left:} first 2 eigenvectors from \dmalg/\LEalg; {\bf middle} after UMAP.  Note that
      UMAP by itself is not able to produce a full-rank embedding
      everywhere; the horseshoe, the two clusters, and the 1
      dimensional ``filament'' between are all artifacts. {\bf Right}: UMAP with selection of the second eigenvector by \cite{yuchaz}. Plots by Yu-Chia Chen.
    }
\end{figure}

\subsubsection{Avoiding the REP in spectral embeddings}
\label{sec:ies-spectral}
For algorithms like \dmalg, and \ltsaalg, the REP has a theoretically straightforward solution. Given a sequence of eigenfunctions 
$F^1,\ldots,F^{m'}\ldots$ on $\M$ (or eigenvectors $\vb^1,\ldots\,\vb^{m'}$ in the finite sample case), with $m'>m$, sorted by their corresponding
eigenvalues, one needs to  select
$F^{j_1}=F^1$, then (recursively) $F^{j_2},\,\ldots F^{j_m}$ so that $\rank
[(\dddF^1)_{\pb},\,\ldots\, (\dddF^{j_m})_{\pb}]=d$ for all $\pb\in\M$. This is called  {\em Independent Eigendirection Selection (IES)}.
In a finite sample, the rank condition must be replaced with the
well-conditioning of  $\dddF$ at the data
points. \cite{Dsilva2018-dz} proposed to measure dependence by regressing
$\vb_{j_{k+1}}$ on the previously selected $\vb_{j_1,\ldots j_k}$; in
\cite{yuchaz}, a condition number derived from the embedding metric
(Section \ref{sec:rmetric}) is used to evaluate entire sets of $m$
eigenvectors.  The {\em manifold deflation} method \cite{tingJordan}
proposes to bypass eigenvector selection by choosing a linear
combination of all eigenvectors that are optimized w.r.t. rank.
Finally, the {\em Low Distortion Local Eigenmaps (LDLE)}~\cite{cloninger} solves the REP by essentially covering the data manifold with contiguous patches (discrete versions of the $U$ neighborhoods) and performing IES on each patch separately. LDLE not only avoids REP, but it is a first step towards the algorithmic use of charts and atlases to complement global embeddings.

\hanyuz{But is it always possible to find $d$ independent eigenvectors? The
answer is positive for the Diffusion Maps algorithm; with $m$ sufficiently large full rank embedding
\citep{berardBessonGallot:94,Portegies:16}\footnote{The result of
\citet{berardBessonGallot:94} is stronger, proving isometry in the
limit.\mmp{mentioned in Isometric emb chapter}} is always
possible. The number $m$ of eigenfunctions needed  may exceed the Whitney embedding dimension and may depend on
injectivity radius, aspect ratio, and so on \citep{Bat14}. We don't have these references yet}

In summary, attraction-repulsion algorithms such as \tsnealg, which are heuristic,
enjoy large popularity due in part to their immunity to the REP,
while eigenvector based methods, although better grounded in theory, are
less useful in practice without post-processing by an IES method.
On the other hand, unlike global search in eigenvector space, a local relaxation algorithm cannot resolve the rank deficiency globally, and it may
become trapped in a local optimum (Figure \ref{fig:ies}).

\subsection{Summary of embedding algorithms}
\label{sec:summ-emb}
A variety of embedding algorithms have been developed. Here we presented representative algorithms of two types. One-shot algorithms that (typically) embed the data by eigenvectors, of which \isomapalg, \dmalg~and \ltsaalg~are the best understood as well as computationally scalable. The main drawback of this class of algorithms is the Repeated Eigendirections Problem, which requires post-processing of the eigenvectors. Neighbor embedding algorithms are (typically) iterative, starting with the output of a one-shot algorithm (\LEalg~for \umapalg) or even PCA. The presence of repulsion makes these algorithms robust to REP which affects one-shot algorithms. Quantifying the repulsion, as well as the smoothness, large sample limits as well as other properties of the neighbor embedding algorithms are less developed at this time. Hence, for the moment, neighbor embedding algorithms remain heuristic for ML, while they remain useful for visualization, and clustering (for which guarantees exist~\cite{lindermanSteinerb:simods19}).

Neither type of algorithm guarantees against local singularities, such as the ``crossing'' in Figure \ref{fig:ies}. Currently, it is not known how these can be reliably detected or avoided. Additionally, all algorithms distort distances except in special cases (as discussed in Section \ref{sec:rmetric}).

All algorithms depend on hyperparameters: intrinsic dimension $d$ (Section \ref{sec:dimension}) or embedding dimension $m$, and $k$ or $r$ for the neighborhood scale (Section \ref{sec:epsilon}).  Iterative algorithms often depend on additional parameters that control the repulsion (such are $\rho$ in \tsnealg), or the descent algorithm.

With respect to computation, constructing the neighborhood graph is
the most expensive step typically for $n$ large. To compound this
problem, finding $k$ or $r$ in a principled way often requires
constructing multiple graphs, one for each scale. One-shot algorithms that compute eigenvectors are quite efficient for $n$ up to $10^6$ when the matrix has a sparsity pattern that corresponds to the neighborhood graph.  Neighbor embedding algorithms work, in theory, with dense matrices (e.g. $\Wb$); however, accelerated approximate versions for these algorithms have been developed such as the Barnes-Hut trees approximation~\cite{vdMaaten:jmlr14}, and the negative sampling heuristic for UMAP~\cite{bohmBehrensKobak:tsne22, mcinnes2018umap}.

\mmp{charts vs. embeddings -- how to phrase it? aren't there border effects from charts?--not now}

\marginpar{
   \includegraphics[width=0.15\textwidth]{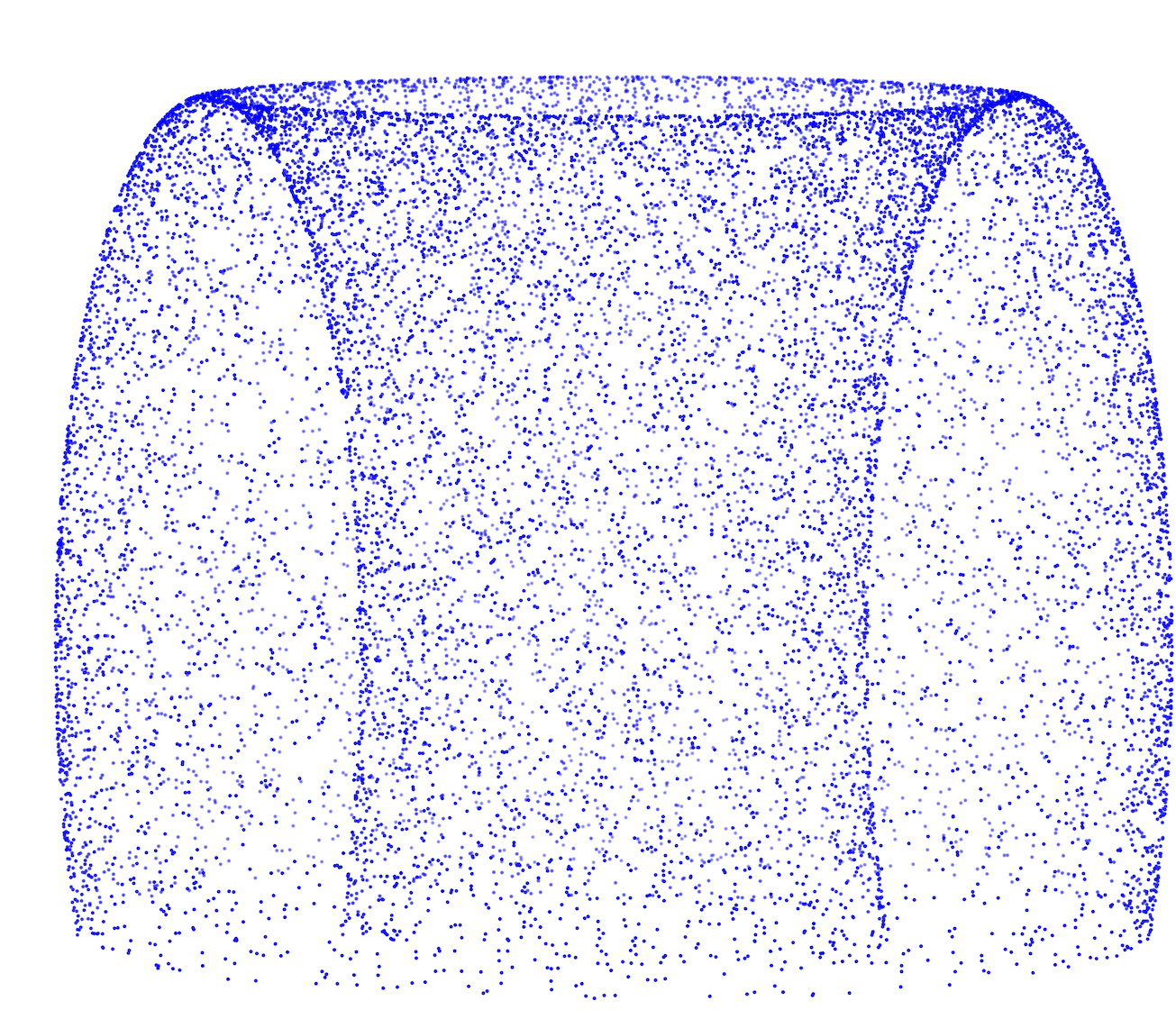}
   \captionof{figure}{Data sampled from the chopped torus}
}

 \begin{figure}
\centering
    \begin{subfigure}[b]{0.3\textwidth}
      \centering
      \caption{Isomap}
      \includegraphics[width=\textwidth]{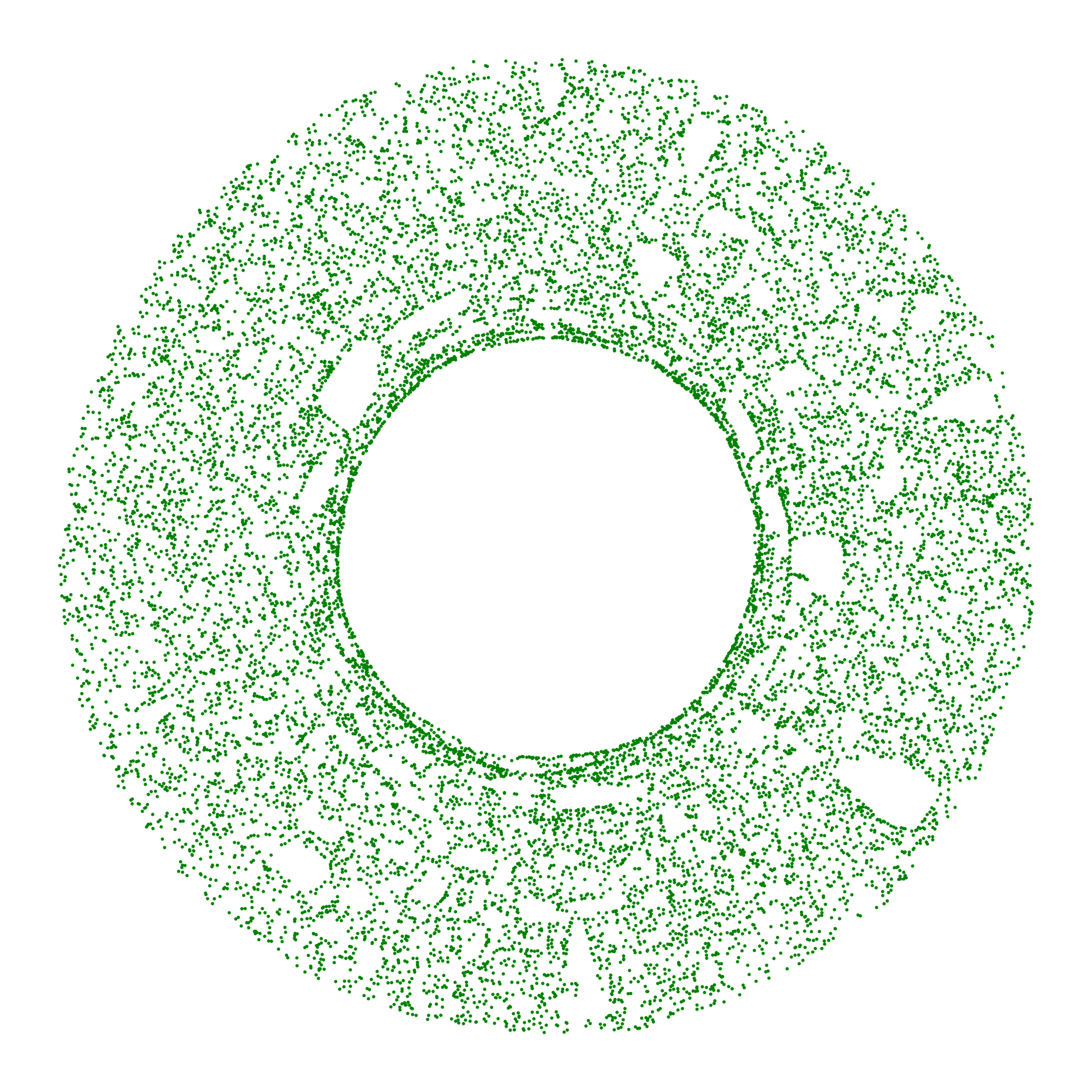}
   \end{subfigure}
    \begin{subfigure}[b]{0.3\textwidth}
       \centering
       \caption{LE}
       \includegraphics[width=\textwidth]{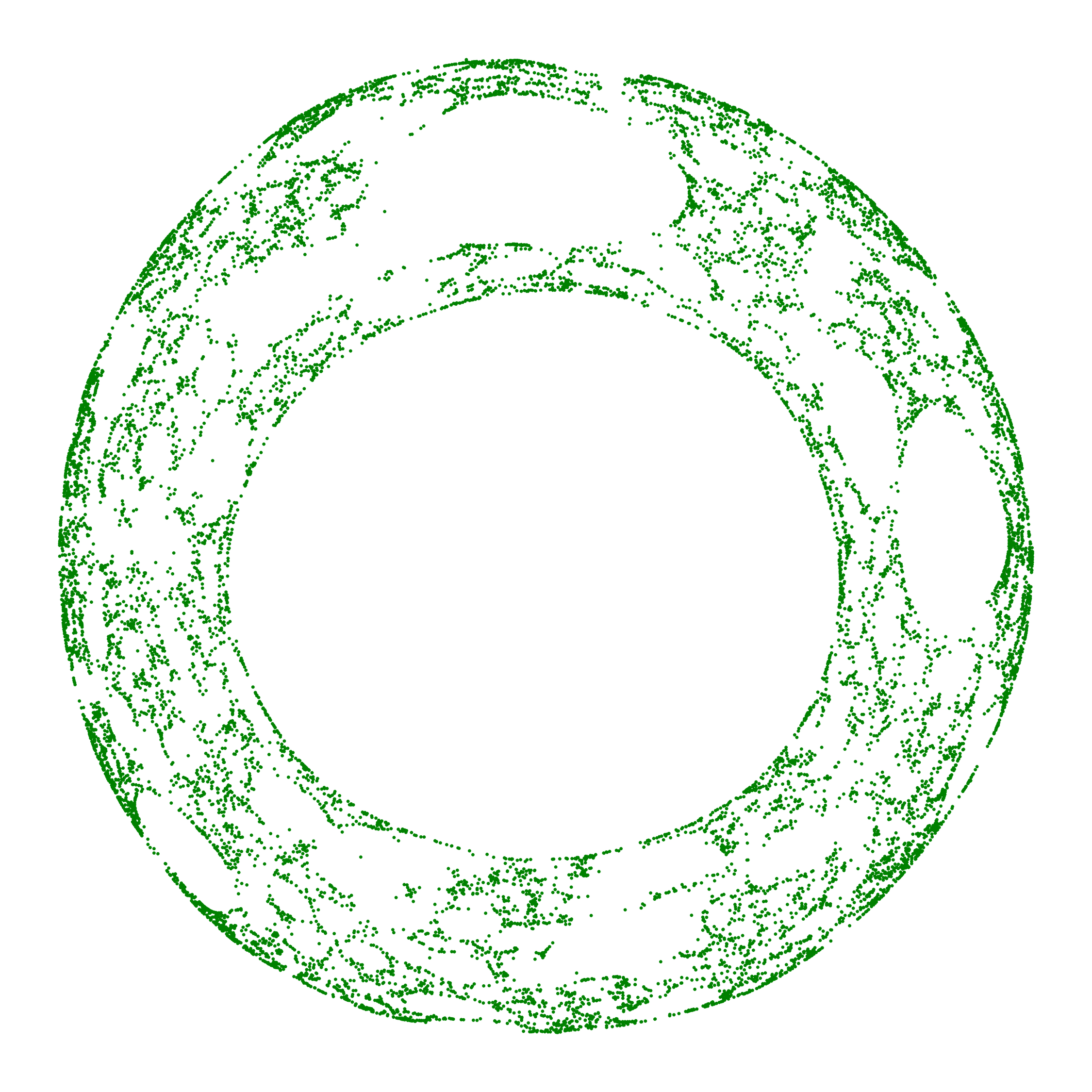}
    \end{subfigure}
    \begin{subfigure}[b]{0.3\textwidth}
       \caption{LLE}
       \centering
       \includegraphics[width=\textwidth]{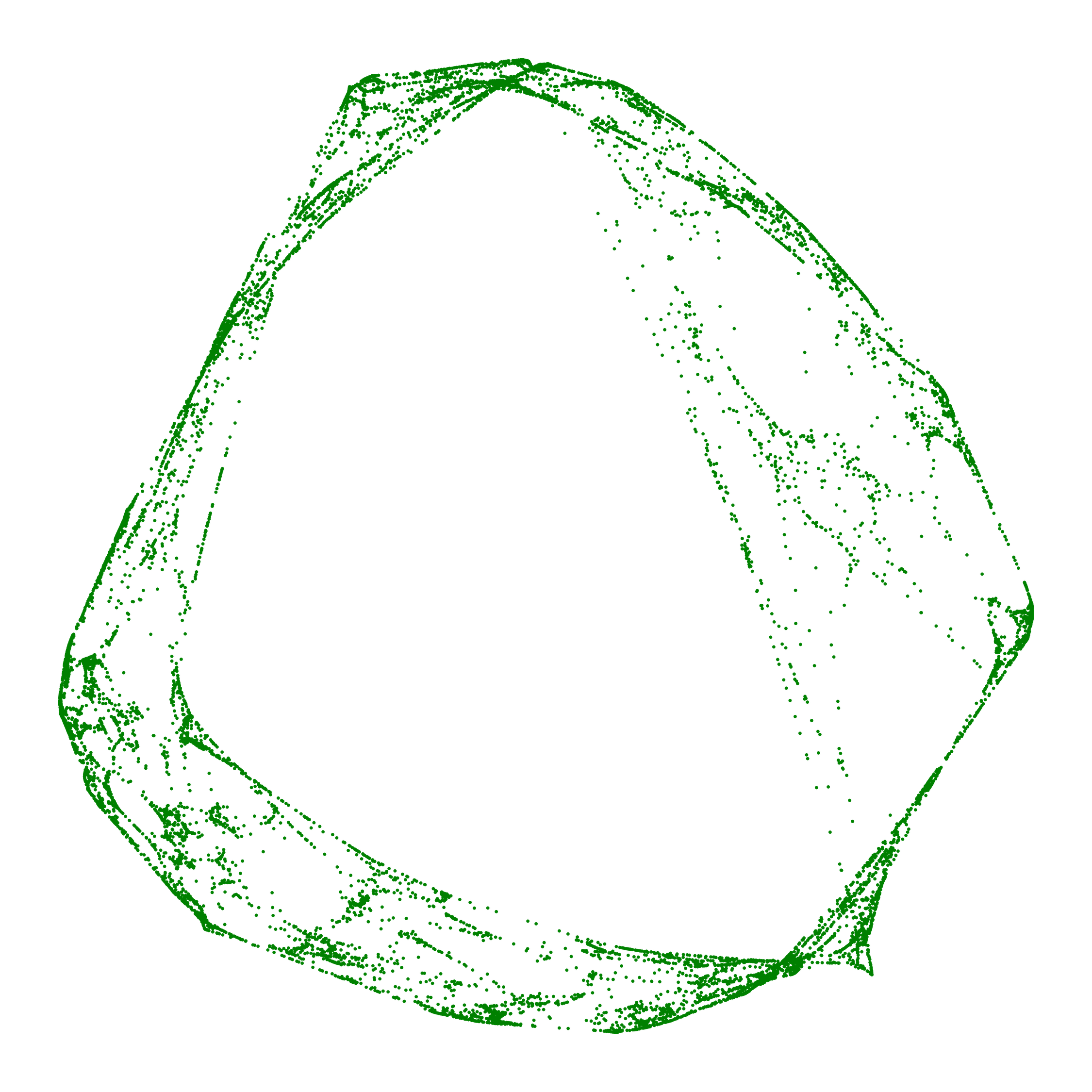}
    \end{subfigure}
    \begin{subfigure}[b]{0.3\textwidth}
       \caption{LTSA}
       \centering
       \includegraphics[width=\textwidth]{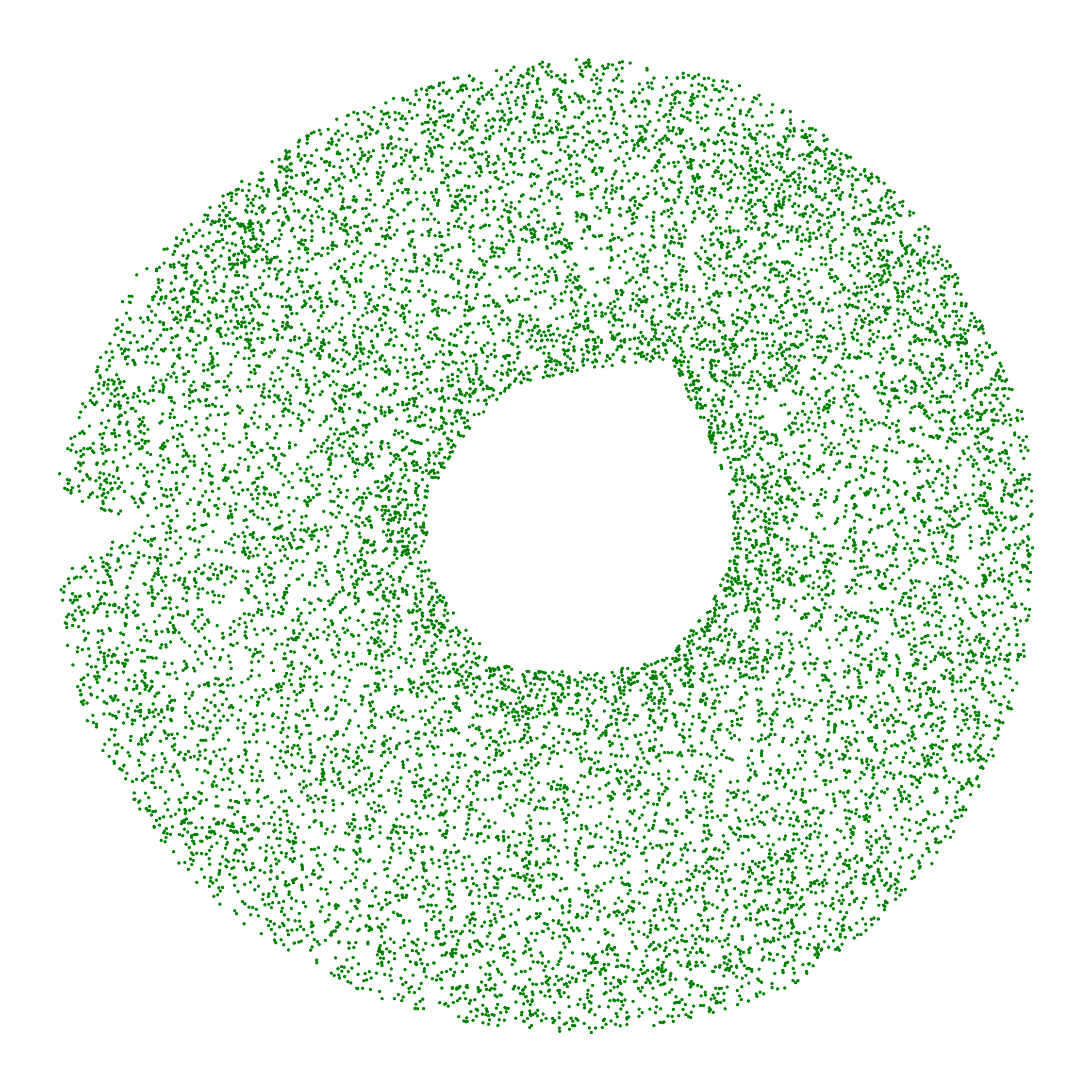}
    \end{subfigure}
    \begin{subfigure}[b]{0.3\textwidth}
       \caption{t-SNE}
       \centering
       \includegraphics[width=\textwidth]{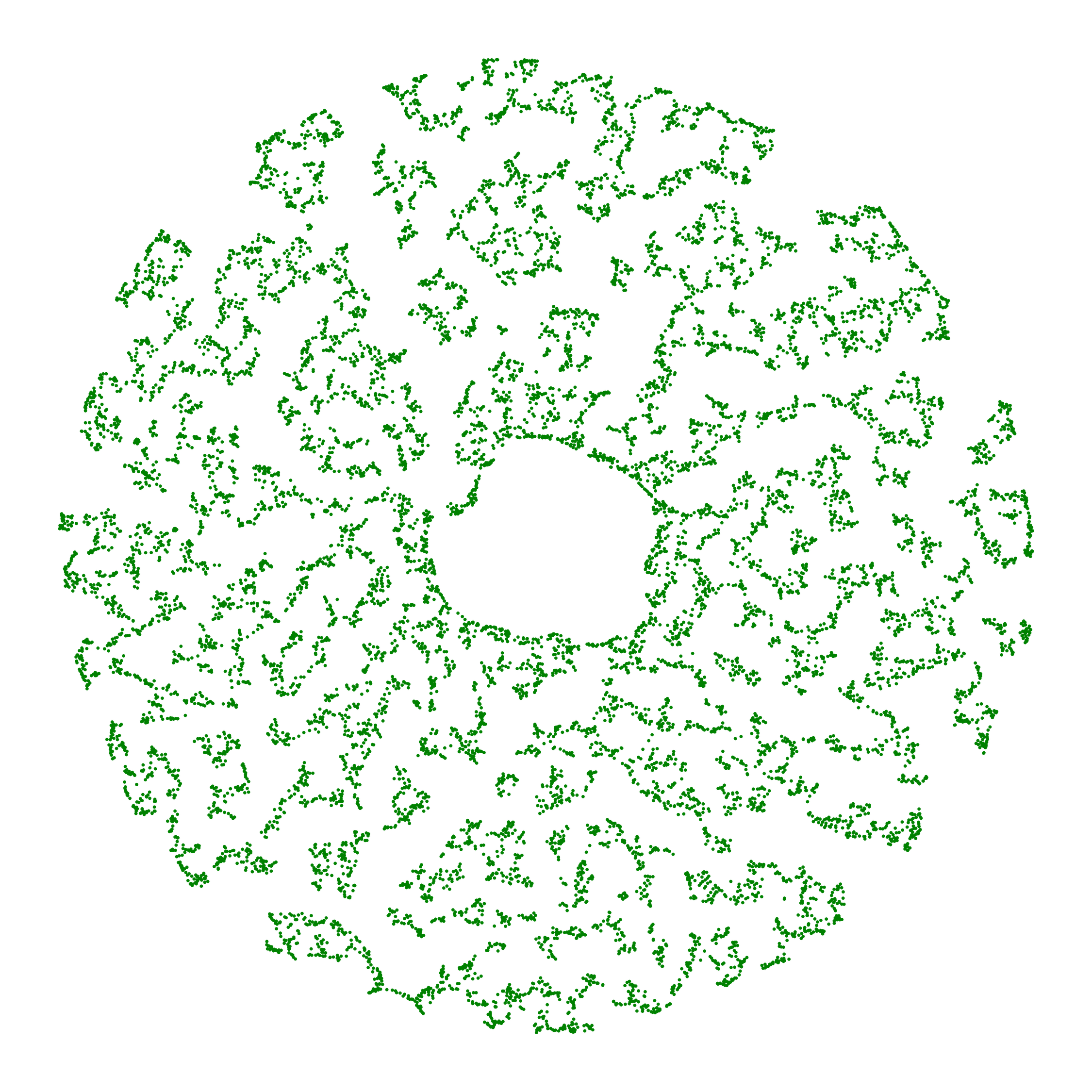}
    \end{subfigure}
    \begin{subfigure}[b]{0.3\textwidth}
      \caption{UMAP}
      \centering
      \includegraphics[width=\textwidth]{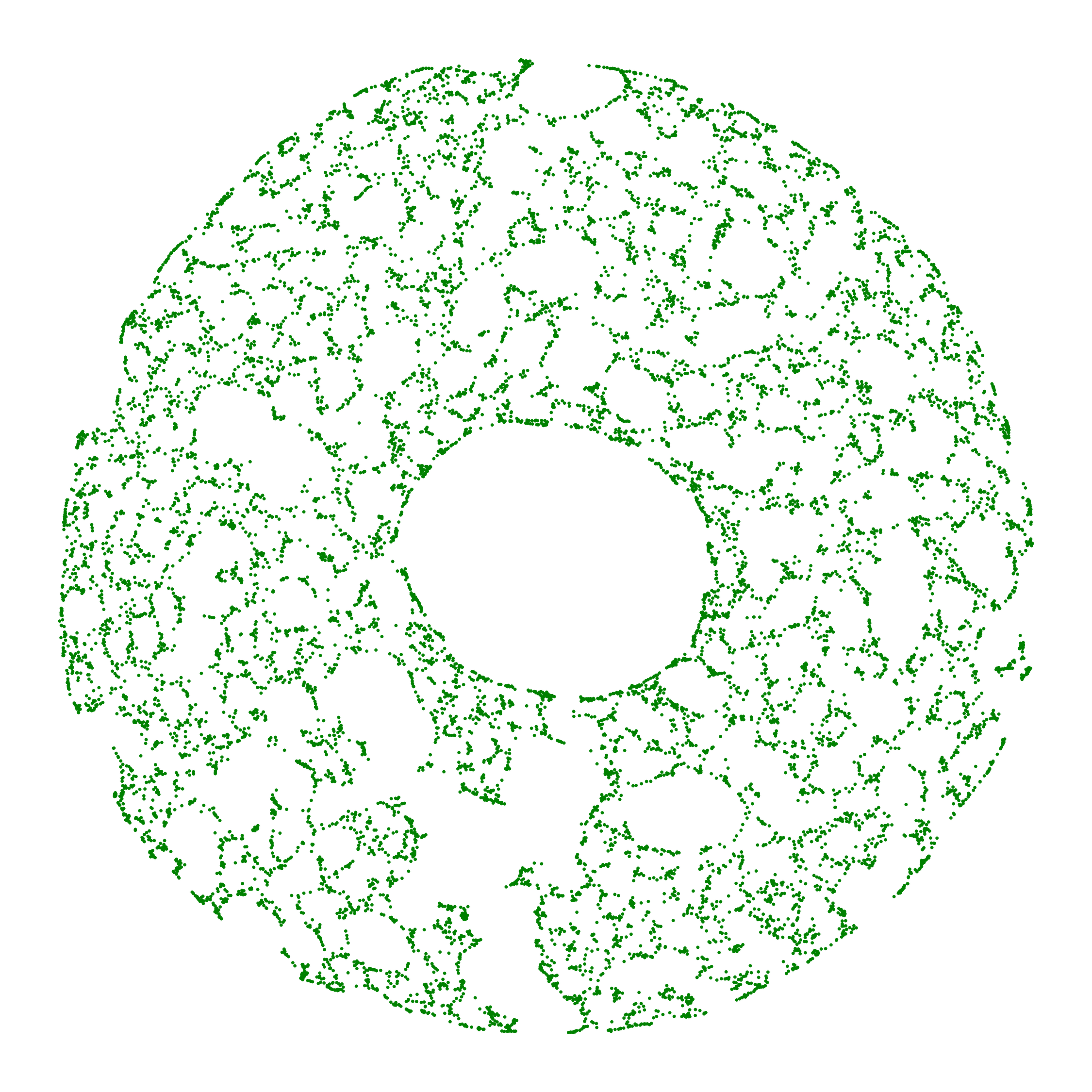}
   \end{subfigure}

    \vspace{0.3em}
    \caption{\label{fig:chopped_torus}Embedding obtained from the algorithms in this section on a chopped torus data set with $n=14,519$ points. \mmp{Specifically, the data are uniformly drawn on part of torus.} This manifold cannot be embedded isometrically in $d=2$ dimensions.
    }
\end{figure}

%% file: rev-geo.tex
\newlength{\backitem}
\setlength{\backitem}{-2em} 
\newcommand{\LB}{\Delta_\M}
The output or result of manifold learning algorithms depends critically on
algorithm parameters such as the type of neighborhood graph (k-nearest
neighbor or radius neighbor), the neighborhood scale ($k$ or $r$), and
embedding dimension $m$ (and intrinsic dimension $d$, in some cases). 

This section is concerned with making these choices in a way that
ensures some type of statistical consistency, whenever
possible. \mmp{say this better} Neglecting statistical consistency and
stats in general is risky. In the worst case, it can lead to methods
that have no limit when $n\rightarrow \infty$ (e.g. for LLE without any regularization), and in milder cases to
biases (e.g. due to variations in data density), and artifacts, i.e.,
features of the embedding such as clusters, arms, and horseshoes that
have no correspondence in the data.

Here we discuss in more general terms what
is known about graph construction methods (Section \ref{sec:graph}),
the neighborhood scale (Section \ref{sec:epsilon}),
and the intrinsic dimension (Section \ref{sec:dimension}). We revisit the estimation of the Laplacian (a normalized version of
the neighborhood graph), as the natural representation of the manifold
geometry, and the basis for the {\sc Diffusion Maps} embedding, which can be seen as
the archetypal embedding (Section \ref{sec:laplacian}).  Finally, in
Section \ref{sec:rmetric}, we turn to mitigate the distortions that
the embedding algorithm currently induces. \mmp{nothing about tangent
space? reach?}

\subsection{Biases in ML. Effects of  sampling density and graph construction}
\label{sec:graph}

\paragraph*{Biases due to non-uniform sampling}
Many embedding algorithms tend to contract regions of $\M$ where the data are densely sampled and to stretch the sparsely sampled regions. In attraction-repulsion algorithms, such as \tsnealg, this is explained by the repulsive forces between every pair of embedding points $\yb_i,\yb_j$, while the attractive forces act only along graph edges. If two dense regions are connected by fewer graph edges, repulsion will push them apart, exaggerating clusters. 

\begin{figure}
\setlength{\mypicwi}{0.28\tttwi}
    \begin{subfigure}[b]{0.3\textwidth} 
    \centering
    \includegraphics[width=\mypicwi]{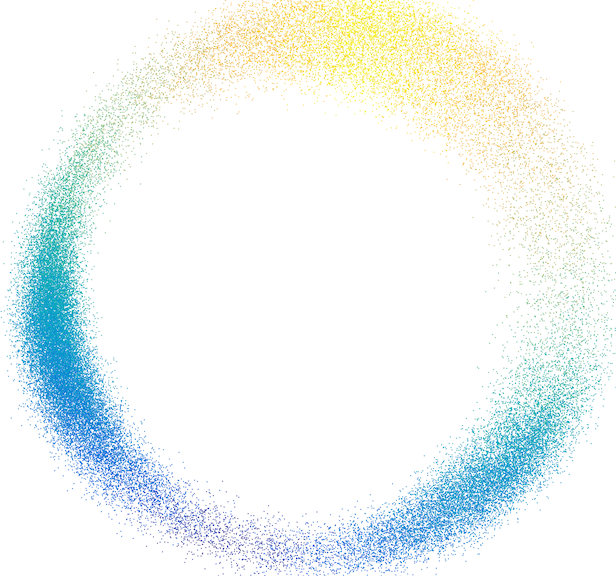}
    \end{subfigure}
    \begin{subfigure}[b]{0.3\textwidth}
       \centering
    \includegraphics[width=\mypicwi]{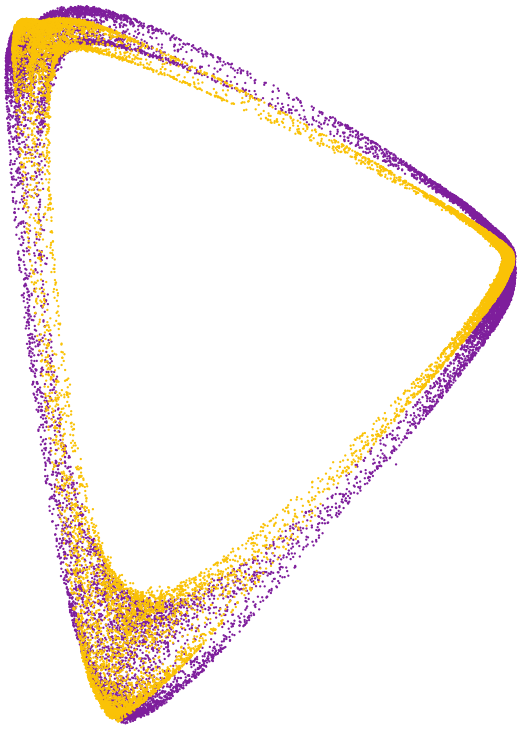}
    \end{subfigure}
    \begin{subfigure}[b]{0.35\textwidth}
       \centering
\includegraphics[width=1\mypicwi]{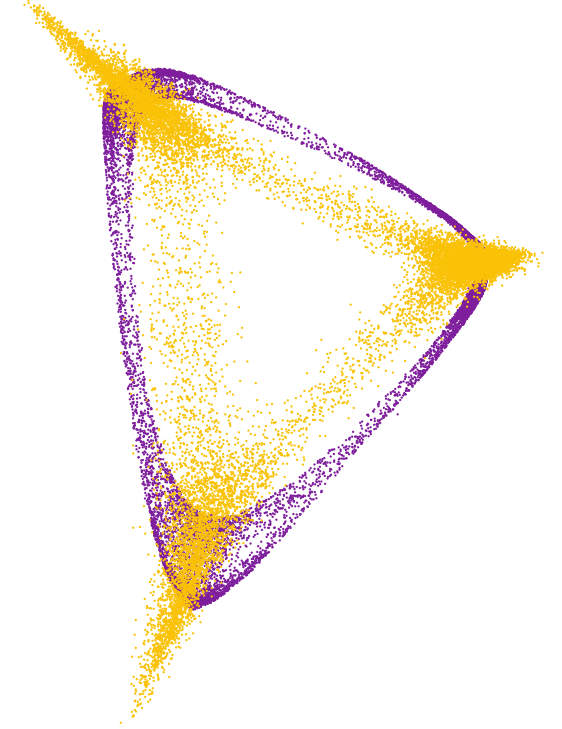}
    \end{subfigure}
\caption{\label{fig:graph-density-effects}
Effects of graph construction and renormalization, when the sampling density is highly non-uniform, exemplified on the configurations of the ethanol molecule. {\bf Left:} original data, after preprocessing, is a noisy torus, with three regions of high density, around local minima of the potential energy. {\bf Center:} Embeddings by \dmalg~(purple), and by the same algorithm with $\Lbw$ constructed from the $k$-nearest neighbor graph (yellow). The low sparse regions are stretched, while the dense regions appear like ``corners'' of the embedding. Note that \dmalg~{\em should} remove the effects of the density; in this case, the variations in density are so extreme that the effect persists. \mmp{i can put a rmetric figure showing it is correctly estimated} The effect is somewhat stronger for the $k$-nearest neighbor graph. {\bf Right:} Embedding by \dmalg~(purple) and by \LEalg~(yellow), which uses the singly normalized $\Lbw^{rw}$.}
\mmp{FYI: I had sized them so that the purple triangle has the same size... no matter, we will have to redo these}
\end{figure}

The effect is similar, albeit less intuitive to explain, for one-shot algorithms, as shown in Figure \ref{fig:graph-density-effects}. For \dmalg~and the graph Laplacian \hanyuz{shall we just keep \dmalg~ here?}, the effect was calculated in \citet{coifman:06}; they also showed that renormalization removes the bias due to non-uniform sampling (asymptotically). Moreover, the degree values $d'_i$ obtained in \lapalg~are estimators of the sampling density around data point $\xb_i$. A simpler method of renormalization, applicable to low dimensional data is to use a simple estimator of the local density, and to use it to renormalize $\Lbw^{rw}$ \cite{yusu09Gradient}. 

If enough samples are available, one can simply resample the data to obtain an approximately uniform distribution. For example, the {\em farthest point heuristic} \mmp{do we have a source} chooses samples sequentially, with the next point being the farthest away from the already chosen points. 

\paragraph*{Effect of neighborhood graph} (Figure \ref{fig:graph-density-effects}) Radius neighbor graph of $k$-nearest neighbors? \mmp{to improve} \cite{TingHJ:10} and later \cite{calderTrillos:19} show that the $k$-nearest neighbor graph, with the similarity matrix with constant kernel $K(u)=1$ exhibits qualitatively similar biases from non-uniform sampling as the simply normalized radius-neighbor graphs \mmp{don' tknow about renormalization +knn ??} 

\mmp{to check LE uses Lrw or Lnorm}\hanyuz{Belkin07 uses Lunorm, Singer06 uses Lrw}

\mmp{limits of algos -- LLE no limit, LTSA etc have limits -- maybe}\hanyuz{After laplacian}

\subsection{Choosing the scale of neighborhood}
\label{sec:epsilon}
\mmp{TODO in this section: scan the refs again, compare rates, find rates for e-vectors? ..., find what manifold quantities are required. In other words, a summary of what rates for what problems with hopefully some lessons learned for the reader. Too technical. Not needed}
\mmp{is it $\epps$ [or $\sqrt{\epps}$]? OK so I guess it's epsilon}
Whatever the task, a manifold learning method requires the user to provide an external parameter, be it the number of neighbors $k$ or the kernel bandwidth $\epps$, that sets the scale of the local neighborhood.

\paragraph*{Theoretical results/Asymptotic results and what they mean}
The asymptotic results of \citet{GinKol06,HeinAL:07,TingHJ:10} and \citet{Singer06} provide the necessary rates of change for $\epps$ with respect to
$n$ to guarantee convergence of the respective estimate. For instance, \citet{Singer06} proves that the optimal bandwidth parameter for Laplacian estimation is given by $h\sim n^{-\frac{1}{d+6}}$ using a random-walk Laplacian. For the $k$-neareast neighbor graph, \citet{calderTrillos:19} show that again for Laplacian estimation, the number of neighbors $k$ must grow slowly with $n$, and a recommended rate is $k\sim n^{\frac{4}{d+4}}(\log n)^{\frac{d}{d+4}}$. The hidden constant factor in these results is not completely known, but they depend on the manifold volume, the curvature, and the {\em
injectivity radius} $\tau$ (typically not known in practice).

With these rate-wise optimal selections of $k$ or $r$, the convergence rate of estimation relating to various objects on the manifold can be established. However, all are {\em
non-parametric} rates. More specifically, they point to the fact that
the sample size $n$ must grow {\em exponentially} with the
dimension. For example, using the previously mentioned rate of $k$, together with the rate of
convergence $\approx \sqrt{\frac{\log
n}{k}}\left(\frac{k}{n}\right)^{1/d}$, one can calculate that, for a
10-fold decrease in error, $n$ must increase by $\approx
10^{(d+4)/3}$. While the actual constants are not
known, the statistical results suggest that, in practice, for one-shot
algorithms, values of $k$ should be sufficiently large, in order to be
close to the maximum accuracy supported by the sample.

For neighbor embedding algorithms, such as \tsnealg, less is known theoretically; however, practically, the defaults are for larger values of $k$, e.g. $k=90$ \cite{Policar19} and some research suggests $k\sim n$, which would create very dense graphs. 

\mmp{allusions to Narayanan and Wasserman in the original paper. to find Wasserman, see if we need to cite them all. To see what the others have for rates}

\paragraph*{Practical methods} Unfortunately, cross-validation (CV), a widely useful model selection method in, e.g., density estimation, is not applicable in manifold learning, for the lack of a criterion to cross-validate. (However, CV is still applicable in {\em semi-supervised} learning on manifolds \cite{belkin-nyogi}.)  The ideas we describe below each mimic CV by choosing a criterion that measures the ``self-consistency'' of an embedding method at a certain scale. 

For the $k$-nearest neighbor graph, \cite{ChenBuja:localMDS09}
 evaluates a given $k$ with respect to the preservation of $k'$
 neighborhoods in the original graph. \mmp{not sure: do they find k'
 in original data and recover neighborhood in embedding, or the other
 way around?}The method is designed to optimize for a specific
 embedding, so the values obtained for $k$ depend on the embedding
 algorithm used. A problem to be aware of with this approach is that
 (see Section \ref{sec:rmetric}) most embeddings distort the data
 geometry, hence Euclidean neighborhoods will not be preserved, even
 at the optimal $k$.

For the radius-neighbor graph, \citet{PerraultM13} seeks to exploit
the connection between manifold geometry, represented by the
Riemannian metric, and the Laplace-Beltrami operator. The radius neighbor graph width $\epps$ affects the Laplacian's
ability to recognize local isometry. Recall that local isometry is
easily obtained by projecting the data on the tangent space at some
point $\xb_i$. The method is specific to the estimation of the
Laplace-Beltrami operator, but in this context, it can be extended to
optimization over other parameters, such as kernel smoothness.

Finally, we mention a dimension estimation algorithm proposed in \citet{Chen2013}, a by-product of this algorithm is a range of scales where the tangent space at a data point is well aligned with the principal
subspace obtained by a local singular value decomposition. As these
are scales at which the manifold looks locally linear, one can
reasonably expect that they are also the correct scales at which to
approximate the manifold.

\subsection{Estimating the intrinsic dimension}
\label{sec:dim}
\input{rev-dimension}

\subsection{Estimating the Laplace-Beltrami operator}
\label{sec:laplacian}
As we discussed in section \ref{sec:algs}, the Laplace-Beltrami operator $\Delta_\M$ serves as an important tool to understand the geometry of a manifold
$\M$. We have seen that the eigenvectors of $\Delta_\M$ can be used to embed the data in low dimensions by the \dmalg~algorithm. Furthermore, if enough eigenvectors are computed, the embedding becomes closer to an isometry \citep{CoiLafLeeMag05}.

Additionally, graph Laplacian estimators of $\Delta_\M$ are used to
measure the smoothness of a function (by
$\frac{1}{2}\bw{f}^T\Lbw \bw{f}$), to provide regularization in
supervised and semi-supervised learning on manifolds \citep{belkin-nyogi,Slep2019}, Bayesian
priors \citep{Kirichenko17LapReg}, or to define Gaussian Processes on a
manifold \citep{Borovitskiy20}.

On one hand, using graph Laplacian to estimate Laplace-Beltrami
operator such as in \citet{coifman:06} has been long
established. Recently, more theoretical results appeared on how this
estimation behaves.
\begin{marginnote}
The \textbf{Laplace-Beltrami operator} $\Delta_{\M}$ acting on a
twice differentiable function $f:\M\rightarrow\rrr$
is defined as $\Delta_{\M}f\equiv$div grad$(f)$.  
\end{marginnote}

In general, two types of convergence have been
studied: \emph{pointwise} and \emph{spectral} convergence under the
formulation of \citet{berry:16}. Let $\mu$ be the Riemannian measure
corresponding to the metric of a $d$ dimensional manifold $\M$, $f\in
C^3(\M)$ be a real-valued function and $q(\bm{x})$ be the sampling
density on $\M$. Further, let
$\bm{f}=(f(\bm{x}_i))_{i=1}^n\in\rrr^n$. Then ideally, the two
convergence paradigm of a random walk Laplacian
$\mathbf{L}^{rw}=\mathbf{D}^{-1}\mathbf{L}^{un}$ defined on an
$\epps$-nearest neighbor graph to its limit $\mathbf{L}^{\infty}$
are given by
\begin{itemize}
    \item Pointwise convergence: $\mathbb{E}[(\mathbf{D}^{-1}\mathbf{L}^{un}\bm{f})_i]\overset{n\rightarrow \infty}{\longrightarrow} c\mathbf{L}^{\infty}f(\bm{x}_i)+O(\epps^2)$
    \item Spectral convergence: $\mathbb{E}[\frac{\bm{f}\mathbf{L}^{un}\bm{f}}{\bm{f}\mathbf{D}\bm{f}}]\overset{n\rightarrow \infty}{\longrightarrow} c\frac{\int_{\bm{x}\in\M}f(\bm{x})(\mathbf{L}^{\infty}f)(\bm{x})q(\bm{x})\mathrm{d}\mu(\bm{x})}{\int_{\bm{x}\in\M} f^2(\bm{x})q(\bm{x})\mathrm{d}\mu(\bm{x})}+O(\epps^2)$; this type of convergence matters for spectral embedding algorithms
\end{itemize}
When the sampling density $q$ is uniform, \citet{belkin:07} showed that pointwise convergence of random-walk Laplacian holds for $\mathbf{L}^{\infty}=\Delta_{\M}$ from an $\epps-$nearest neighbor graph. \citet{coifman:06} showed that pointwise convergence holds for $\mathbf{L}^{\infty}=\Delta_{\M}-\Delta_{\M} q/q$ when $q$ is not uniform. Through the renormalization, $\Lbw$ as in algorithm \ref{alg:renormLap} can eliminate the bias term $\Delta q/q$ and converge to Laplace-Beltrami operator regardless of sampling density. \citet{TingHJ:10} further showed that for $K-$nearest neighbor graph, the random-walk graph Laplacian pointwisely converge to $\Delta_{\M}$ rescaled with $q^{2/d}$. For spectral convergence, readers are encouraged to consult \citet{belkin:07,berrysauer16,GarciaTrillos2018,GarciaTrillos2020}.

\mmp{limits of other algorithms will here or next sections?? from Ting: LTSA, LLE, and more generally about limits of Laplacians?}

Recently, the limits of a class of manifold learning algorithms to differential operators are studied. For a specific type called linear smoothing algorithms, these ML algorithms are proved to converge to a second-order differential operator on $\M$. For example, \LEalg~,\dmalg~ converges to Laplacian operator, {\sc LTSA},{\sc Hessian Eigemaps} both converge to the Frobenius norm of Hessian. Unregularized {\sc LLE}, on the other hand, fails to converge to any differential operator. Details can be found in \citet{Ting2018}. 

\subsection{Embedding distortions. Is isometric embedding possible?}
\label{sec:rmetric}
Figure \ref{fig:chopped_torus} shows the outputs of various embedding algorithms on
a simple 2-manifold $\M\subset \rrr^3$. It is easily seen that the
results depend on the algorithm (and parameter choices), as well as on
the input (manifold and sampling density on $\M$). While all are smooth embeddings the algorithm-dependent distortions -- amounting to different coordinate systems -- make these outputs irreproducible and incomparable.

The presence of distortion is commonly observed empirically.
Note that the distortions {\em do not disappear} when the sample size $n$
increases or the sampling density is uniform, or even when the
consistent graph and Laplacian are used. They are also not an effect of
sampling noise. 
This section is concerned with recovering reproducibility, by
preserving the intrinsic geometry of the data. 

\paragraph*{Attempts at isometric embedding}
\label{sec:geo-isometric}
Mathematically, the presence of distortions means that an embedding
$F$ is not isometric. Distortionless, i.e. isometric embedding is
possible, as proved by a famous result of John Nash (Nash embedding theorem, \citet{smoothmfd}). Note that
for a smooth embedding, the number of dimensions required is $m\geq
d(d+1)/2$, the number of degrees of freedom of $\g$. \mmp{In the special
case of submanifolds of $\rrr^D$, who inherit their metric from
$\rrr^D$, $m$ is trivially no larger than $D$.} The proof of Nash's
theorem is constructive, but not easily amenable to consistent,
numerically stable implementation. 

A more recent seminal result is that the \dmalg~embedding is isometric
for large $m$ (e.g. $m\rightarrow \infty$)
\citep{berardBessonGallot:94,Portegies:16}\mmp{also $t\rightarrow 0$ how shall we
express it?}. While these results are important mathematically, the
fact that the embedding dimension $m$ is required to be large makes
them less interesting for data scientists/defeats the goal of
dimension reduction. \mmp{verma here?}

Many ML methods focus on promoting isometry in local
neighborhoods. Apart from the previously mentioned {\sc Hessian Eigenmaps} \citep{Donoho2003HE}, {\sc LTSA}\citep{ZhangZ:04}, the method of \citet{weinbergerSaul:MVU06} preserve local distances in a
  Semidefinite Programming (SDP) framework. {\sc Conformal Eigenmap} in \citet{sha:05} maps triangles in each neighborhood, thus succeeding in preserving angles. The works of
  \citet{tong10} and \citet{linHeetal:jmlr13} approach global
  isometry by means of constructing \myemph{normal coordinates}
  recursively from a point $\pb\in\M$, or, respectively, by mutually
  orthogonal \myemph{parallel vector fields}, and \citet{verma} is
  the first attempt to implement Nash's construction. 
  We note that, with the exception of \citet{verma}, these methods do not guarantee an isometric embedding except in limited special cases. 
\mmp{DON'T KEEP Note that other paradigms are also prone to biases; for instance PCS -- bias persists in noise, due to KDE even when no noise (but decays to 0). How about lPCA?}

\paragraph*{Preserving isometry by estimating local distortion}
\label{sec:rmetric-alg}
While finding a practical isometric embedding algorithm has been unsuccessful so far, removing the distortions is possible for any well-behaved embedding algorithm by a post-processing approach \citep{PerraultMMcQueen:nips17}. The idea is simple and general: given the (distorted) output $\yb_1,\ldots \yb_n$ of an embedding algorithm on data $\xb_1,\ldots \xb_n$, one can \myemph{estimate} the distortion incurred at each point. Once the distortions are known, whenever a distance, angle, or volume is calculated, one applies local corrections that amount to obtaining the same result as if the embedding was isometric.

This is always possible via the \mydef{push-forward metric}. Let
$(\M,\g)$ be a Riemannian manifold, and $F:\M\rightarrow
\N=F(\M)\subset \rrr^m$ a smooth map, representing, e.g., the limit
case of an embedding algorithm.  We can endow $\N$ with the
\mydef{push-forward Riemannian metric} $\gpush$ of $F$ at point
$\pb\in\mathcal{M}$. Let $\ub,\vb\in\T_{F(p)}\N$ be vectors in the
tangent space of $\N$ at point $F(\pb)$. Then the push-forward of
$\g$ at $\pb$ is defined by 
\begin{eqnarray}
\label{eq:push-f}
\langle \ub,\vb\rangle_{\gpush(F_{\pb})}
& \equiv & \langle
  \dddF_{\pb}^{\dagger}\left(\ub\right),\dddF_{\pb}^{\dagger}\left(\vb\right)\rangle_{\g(\pb)}\,. \end{eqnarray}
In the above, $\dddF_{\pb}^{\dagger}$ is the pseudoinverse of   $\dddF_{\pb}$. In matrix notation \eqref{eq:push-f}
implies that 
\beq \label{eq:push-f-mat}
\gpush(F_{\pb}) \;\equiv\;((\dddF_{\pb})^{T})^{\dagger}\g(\pb)\left(\dddF_{\pb}\right)^{\dagger}
\eeq
with $\dddF_{\pb},\g(\pb),\gpush(\pb)$ are matrices of size $d\times
m,\,d\times d$ and $m\times m$ respectively, and $\g(\pb),\gpush(\pb)$
positive semidefinite matrices of rank $d$.\mmp{Should this be in the background?}
 When $\mathcal{M}\subset\rrr^{D}$, with metric inherited from the
ambient space, $\g(\pb)=\mathbf{I}_d$ the unit matrix and
$\gpush(\pb)=((\dddF_{\pb})^{T})^{\dagger}(\dddF_{\pb})^{\dagger}$.
\mmp{COMPRESS? When $\mathcal{M}\subset\rrr^{D}$, with metric inherited from the
ambient Euclidean space, as is often the case for manifold learning,
$\g$ is simply the Euclidean metric in $\rrr^{D}$ projected onto
$\tp$. From the above definition, it follows that $\gpush$ is
symmetric semi-positive definite of rank $d$, positive definite on
$\T_{\pb}F(\M)$ and null on $\T_{\pb}F(\M)^{\perp}$.}
Comparing \eqref{eq:push-f} with \eqref{eq:defi-isometry} it is easy
to see that $(F(\M),\gpush)$ is isometric with the original
$(\M,\g)$. 

Hence, if one computes for each embedding point $\yb_i$ the respective pushforward metric $\gpush_i\in \rrr^{m\times m}$, then all geometric quantities computed with the points $\yb_1,\ldots \yb_n$ w.r.t. $\gpush$ would preserve their values in the original data, subject only to sampling noise.

It remains to see how to estimate $\gpush$. A direct way is via \eqref{eq:push-f-mat}, using an estimator of $\dddF(\bm{p})$. \mmp{do we know of one?}
Another method~\cite{PerraultM13} is via the Laplace-Beltrami operator $\Delta_{\M}$, namely using the Diffusion Maps Laplacian, whose properties and
consistency is well studied, as seen in Section \ref{sec:laplacian}.
To extract $\gpush$, \cite{PerraultM13} applies $\Delta_\M$ to a
suitably chosen set of \myemph{test functions} $f_{kl}$, with $1\leq
k\leq l\leq m$, where
$f_{kl,\pb}=\left(F^{k}-F^{k}(\pb)\right)\left(F^{l}-F^{l}(\pb)\right)$
are pairwise products of coordinate functions, centered at point
$\pb$. They show that
$\frac{1}{2}\Delta_{\M}f_{kl,\pb}|_{(\pb)}=\ginv_{kl}(\pb)$,
the $k,l$ entry in the inverse pushforward metric at
$\pb$ (algorithmically, on a sample, this operation can be easily vectorized).  To obtain $\g_i$ the metric at data point $i$, one computes
the rank $d$ pseudo-inverse~\citep{BenGre03} of $\ginv_i$ by SVD. Note
that $\ginv_i$ itself measures the local distortion at data point $i$. 
 \mmp{Figure comments on example}  \mmp{we should include this \citep{rosenberg_1997}}
The embedding metric $\gpush$ and its SVD offer other
insights into the embedding. For instance, the singular values of $\gpush$
may offer a window into estimating $d$ by looking for a ``singular
value gap''. The $d$ singular vectors form an orthonormal basis of the
tangent space $\T_{\pb}F(\M)$ at point $\pb=\yb_i$, providing a
natural framework for constructing a \mydef{normal} coordinate chart
around $\pb$. The non-zero singular values of $\ginv_i$ yield a
measure of the distortion induced by the embedding around the data point $\xb_i$ (indeed, if the embedding were isometric to $\M$ with the metric
inherited from $\rrr^m$, then the embedding metric $\gpush$ would have
exactly $d$ singular values equal to 1).

This last remark can be used in many ways, such as getting a global
distortion for the embedding, and hence as a tool to compare various
embeddings. It can also be used to define an objective function to
minimize in order to get a more isometric embedding; such as the
\rralg~of \citet{PerraultMMcQueen:nips17}. 

\begin{figure}[!b]
  \centering
      \begin{subfigure}[b]{0.3\textwidth}
        \centering
        \caption{Isomap}
        \includegraphics[width=\textwidth]{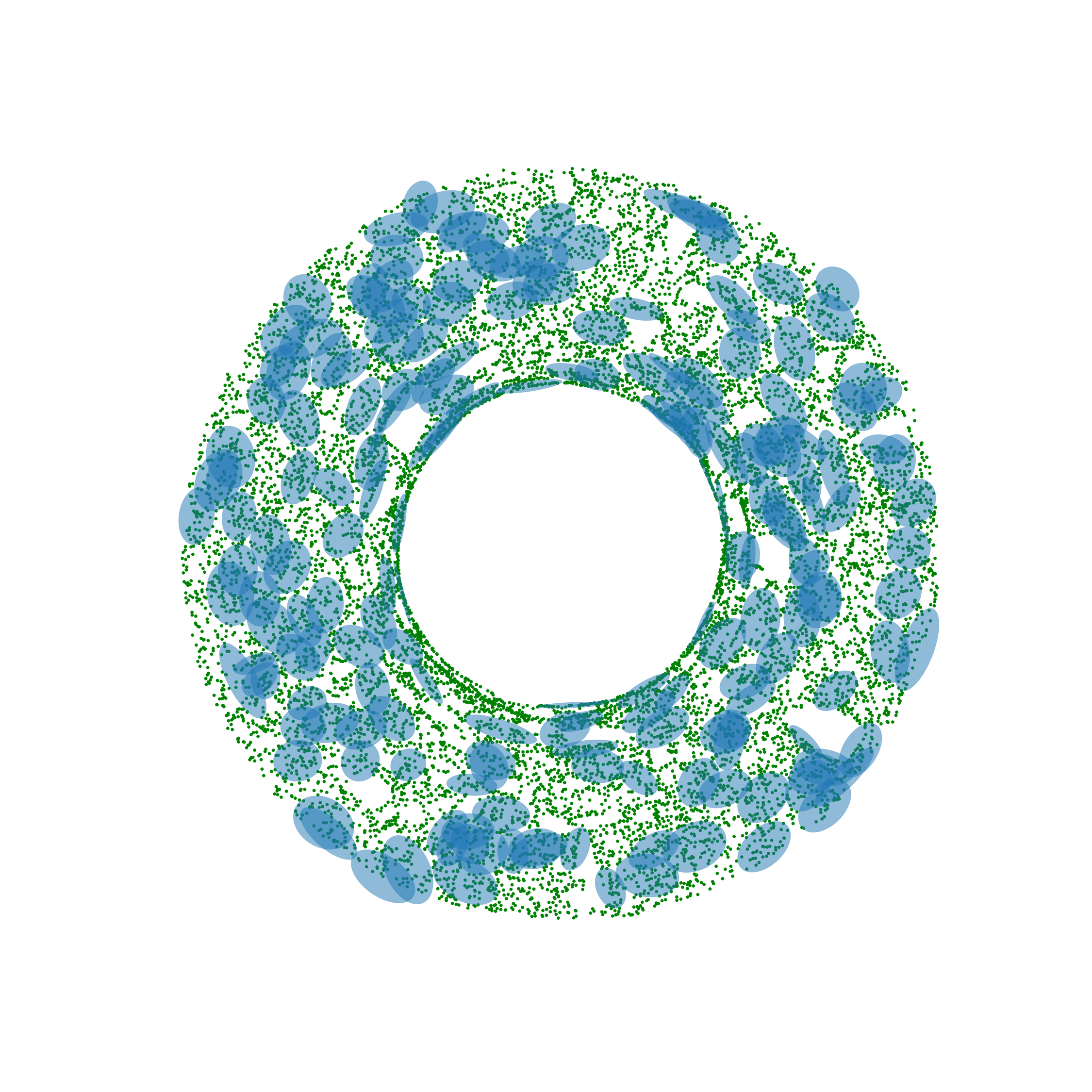}
     \end{subfigure}
      \begin{subfigure}[b]{0.3\textwidth}
         \centering
         \caption{LE}
         \includegraphics[width=\textwidth]{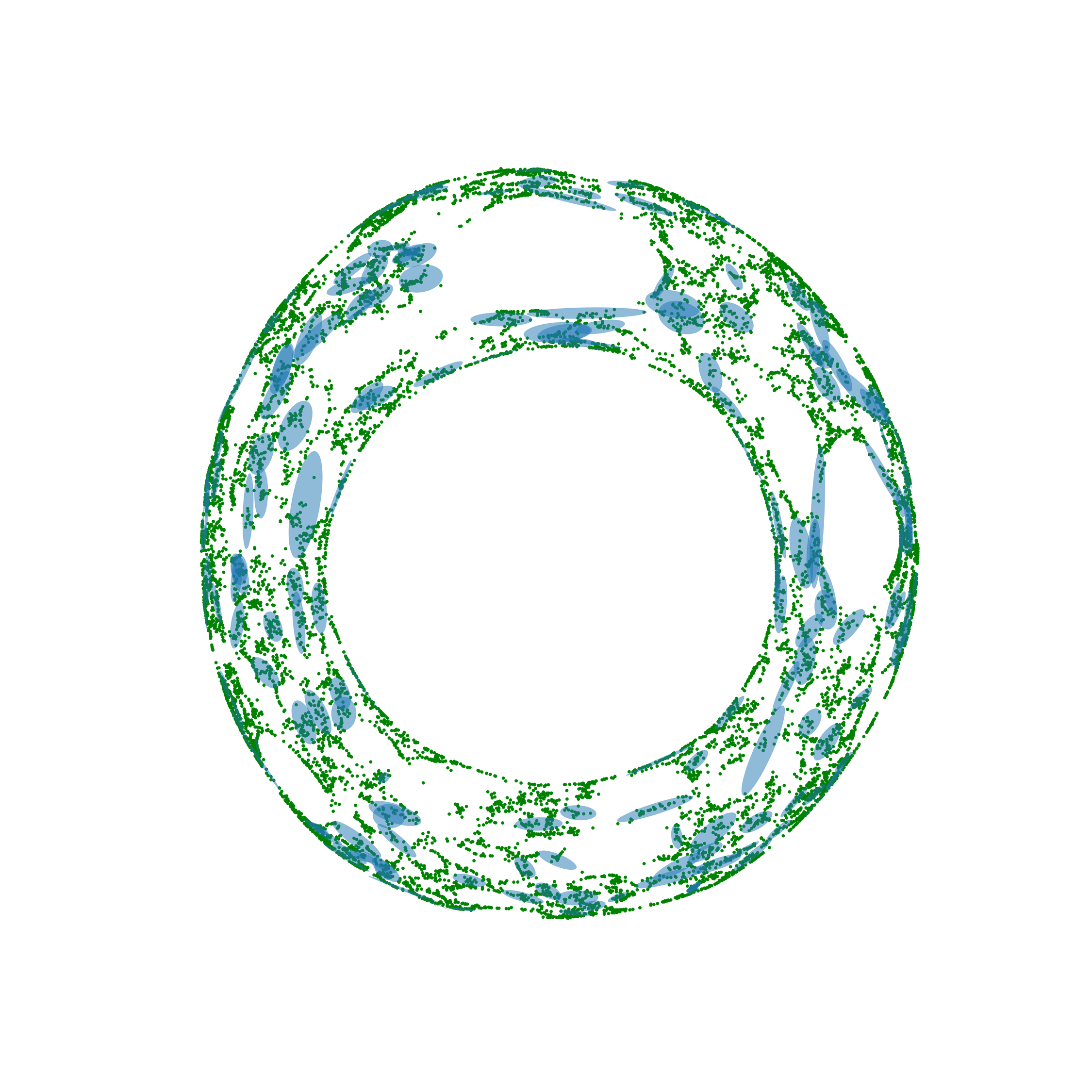}
      \end{subfigure}
      \begin{subfigure}[b]{0.3\textwidth}
         \caption{LLE}
         \centering
         \includegraphics[width=\textwidth]{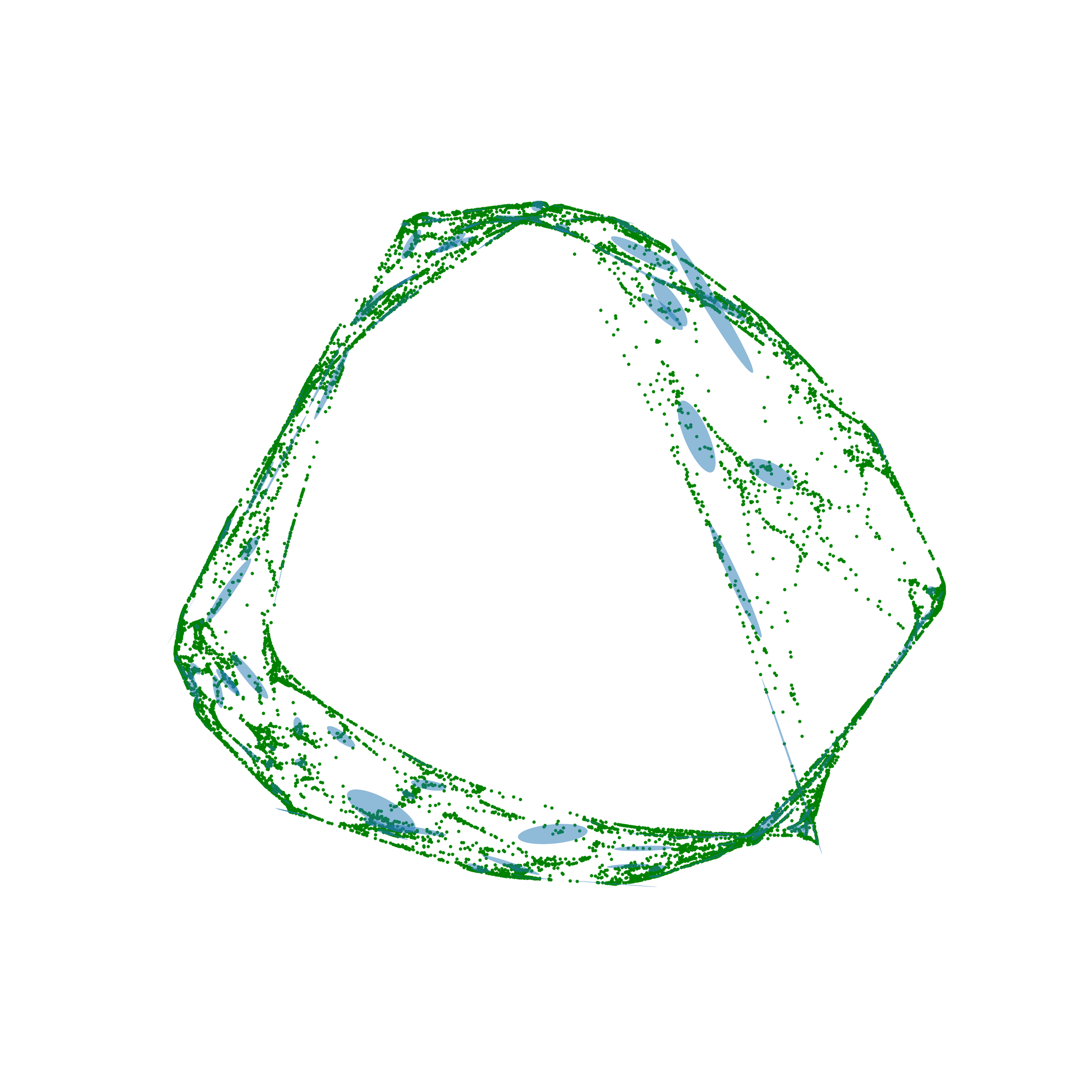}
      \end{subfigure}
      \begin{subfigure}[b]{0.3\textwidth}
         \caption{LTSA}
         \centering
         \includegraphics[width=\textwidth]{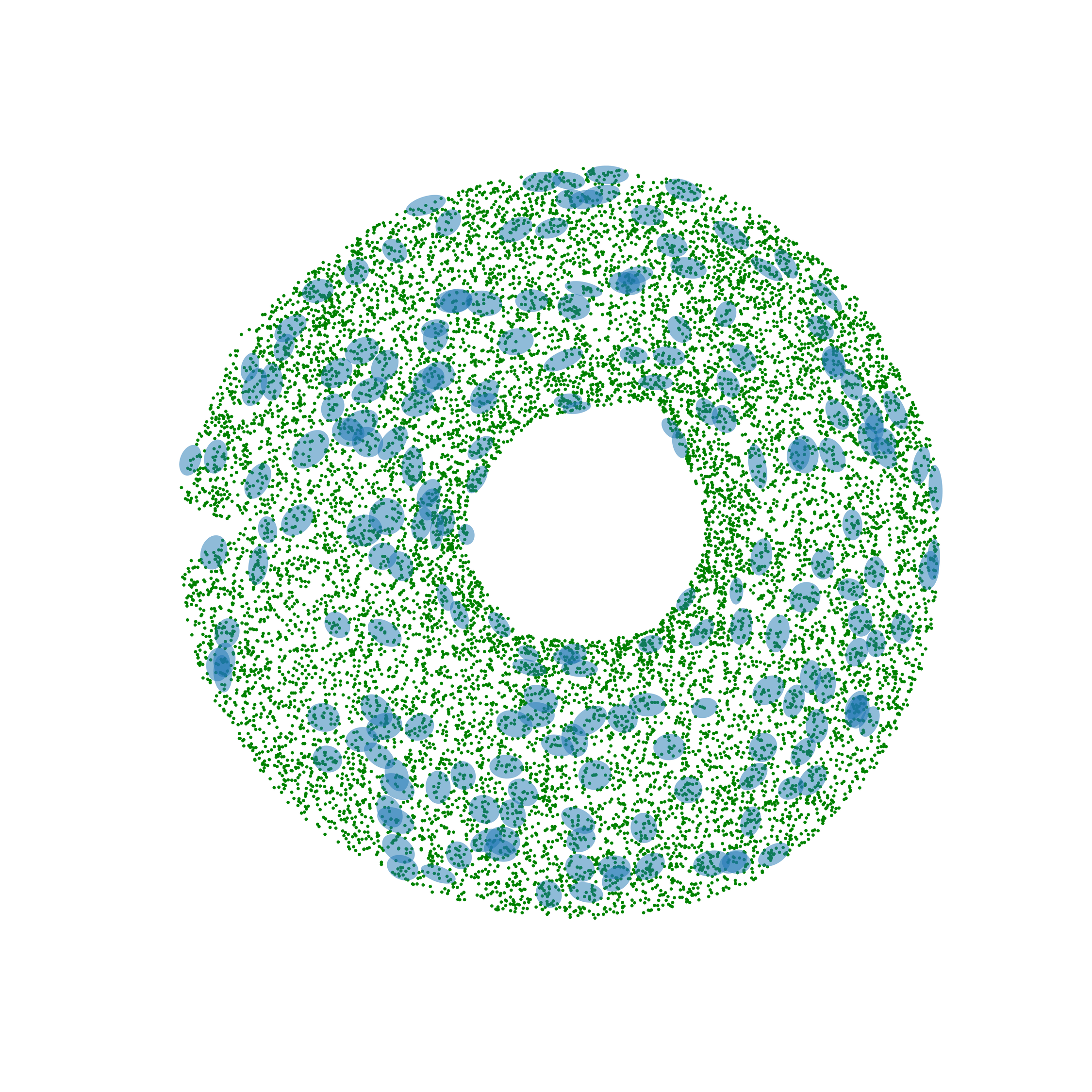}
      \end{subfigure}
      \begin{subfigure}[b]{0.3\textwidth}
         \caption{t-SNE}
         \centering
         \includegraphics[width=\textwidth]{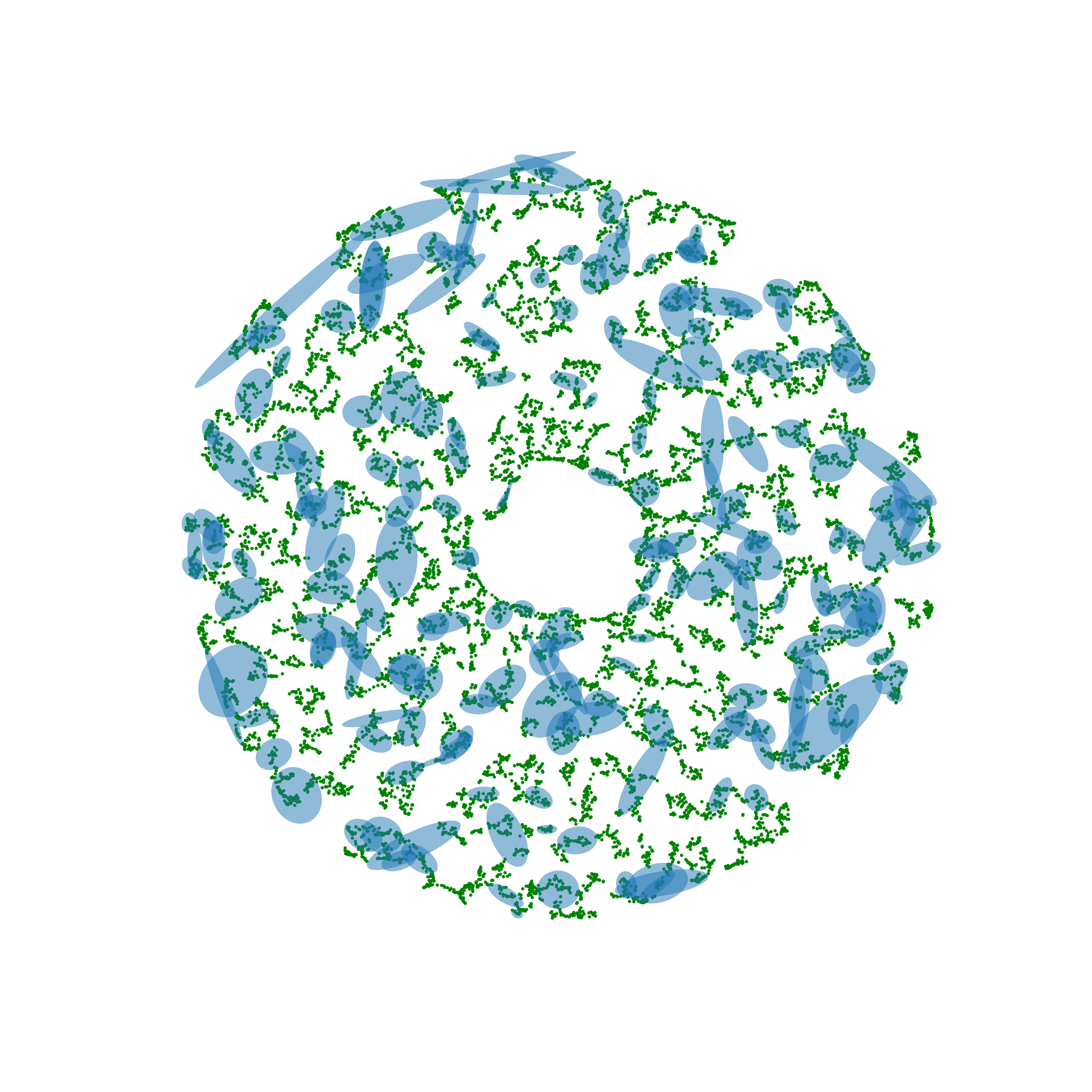}
      \end{subfigure}
      \begin{subfigure}[b]{0.3\textwidth}
        \caption{UMAP}
        \centering
        \includegraphics[width=\textwidth]{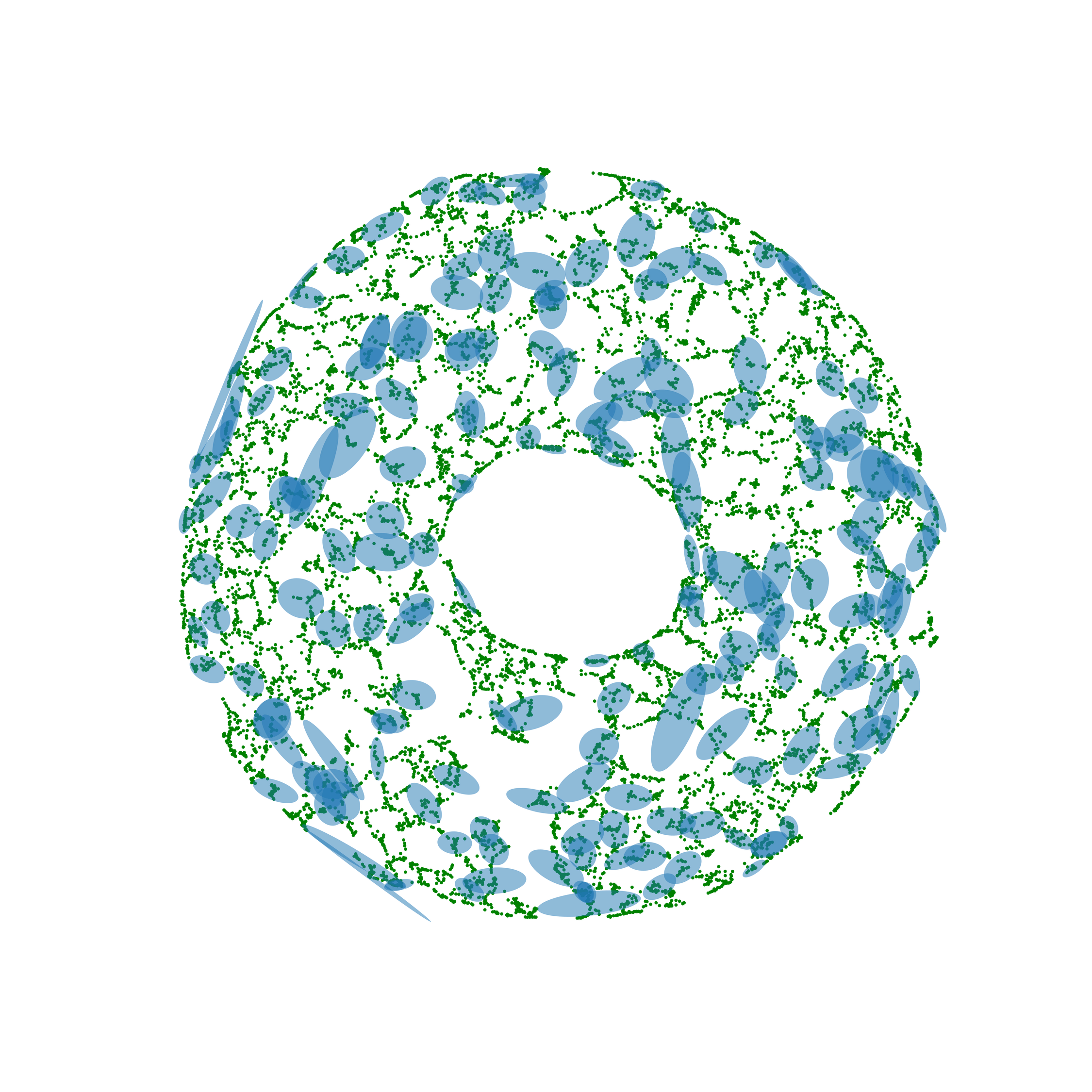}
     \end{subfigure}
  
      \vspace{0.3em}
      \caption{\label{fig:chopped_torus_rmetric}The embeddings from Figure \ref{fig:chopped_torus}, with the distortion $\ginv$ estimated at a random subset of points.
      }
  \end{figure}

%% file: rev-dimension.tex
\label{sec:dimension}
Knowing the intrinsic dimension of data is important in itself. Additionally, some embedding algorithms ({\sc Isomap, LTSA}), as well as all local PCA and Principal $d$-manifolds algorithms require the intrinsic dimension $d$ as input. 

\paragraph*{How hard is dimension estimation?}
The dimension of a manifold is a non-negative integer, therefore, intuitively, it should require fewer samples to estimate than a real-valued geometric parameter. Indeed, it is known \citet{jisukim19, GenovesePVW12} that the \mydef{minimax rate} for dimension estimation are between $n^{-2n}$ and $n^{-\frac{n}{D+1}}$ for a well-behaved manifold $\M$. This is an information-theoretic result, delimiting what is possible: the minimax rate is the best possible rate for any dimension estimator, on the worst possible distribution for this estimator. In the case when i.i.d. noise \mmp{but noise known}is added to the sample, \citet{koltchinskii:00} shows that the minimax rate is exponential, i.e. of order $q^n$ for some $q<1$. This rate is the probability that $\hatdim\neq d$; \citet{koltchinskii:00} also proposes an estimator. 

Unfortunately, the empirical experience belies the optimistic theoretical results. Due primarily to the presence of noise, which does not conform to the above assumptions, and secondarily to non-uniform sampling, estimating $d$ is a hard problem, of which no satisfactorily robust solutions have been found yet (see \citet{solla} for some empirical results). 

\paragraph*{Principles and methods for estimating $d$}
\mmp{keep this? One of the earliest algorithms is that of \citet{fukunaga} PCA.}
\mmp{all these are the same for the manifold of dimension d}
An idea that appears in various forms through the dimension estimation literature is to find a {\em local statistic} that scales with $d$ by a known law. For example, the volume of a ball $B_r$ of radius $r$ contained in a manifold $\M$ is proportional to $r^d$. If we take $n$ samples uniformly from $\M$, the number of samples contained in $B_r$, denoted $\#B_r$ is proportional to $nr^d$, or equivalently
\beq
\log \#B_r\;=\;d\log r+\log n +\text{constant}.
\eeq
This suggests if we fit a line to $(\log r, \log \#B_r)$, the slope of the line would represent $d$.

Recall that $k_{i,r}$ represents the number of radius $r$ neighbors of data point $x_i$. Hence $\log k_{i,r}\approx d\log r+ \text{constant}$. This is the idea of \citet{grassberger1983} who introduced the \mydef{correlation dimension} estimator given by 
\beq \label{eq:dim:correl-dim}
\hatdim_C \;=\;\lim_{r\rightarrow 0}\frac{\log \frac{1}{n}\sum_{i\neq i'}\onevector_{\|x_i-x_{i'}\|\leq r}}{\log r}
\eeq
In the above, $\frac{1}{n}\sum_{i\neq i'}\onevector_{\|x_i-x_{i'}\|\leq r}$, where the sum is taken over unordered pairs, is nothing else but $\frac{1}{2n}\sum_{i=1}^n (k_{ir}-1)$; hence, the correlation dimension uses an average number of neighbors. This estimator is easily computed for the radius neighborhood graph. To sidestep the inconvenient assumption that the sample is uniform over $\M$, other methods consider statistics such as $\frac{k_{i,2r}}{k_{i,r}}\approx 2^d$, which lead to so-called \mydef{doubling dimensions}. \citep{Assouad1983}

Similarly, the covering number $\nu(r)$, representing the minimum number of boxes (in this instance) of size $r$ needed to cover a manifold, scales like $r^{-d}$ for $r$ small. The \mydef{Box Counting dimension} \citep{Falconer03-boxcounting} of an object is defined as
\beq
\label{eq:dim:box-dim}
\hatdim_{BC}\;=\;\lim_{r\rightarrow 0}\frac{\ln \nu(r)}{\ln\frac{1}{r}}.
\eeq
If $\nu(r)$ is defined by way of balls \mmp{do this first?} the above becomes the well-known \mydef{Haussdorff dimension} \citep{Falconer03-hausdorff}. When $\hatdim_{BC}$ is estimated from data, the covering number represents the number of boxes (balls) to cover the data set. 
Note that for finite $n$, $r$ cannot become too small, as in this case, every ball or box will contain a single point. The finite radius $r$ is a scale parameter trading off bias (which increases with $r$), and variance (which decreases with  $r$). \mmp{find some sources for choice of r/ say we'll discuss later}

All the above estimates converge to the intrinsic dimension $d$ when the data is sampled from a $d$-manifold. In practice, and with noise, their properties differ, as well as the amount of computation they need.

Modern estimators consider other statistics, such as distance to $k$-th nearest neighbor \citep{pettis79,costa05}, the volume of a spherical cap \citep{Kleindessner2015DimensionalityEW} (both statistics can be computed without knowing actual distances, just comparisons between them), or Wasserstein distance between two samples of size $n$ on $\M$, which scales like $n^{-1/d}$\citep{Block22}; the algorithm of \citet{LevinaB04}, analyzed in \citet{farahmand07}, proposes a Maximum Likelihood method based on $k$-nearest neighbor graphs. 

\mmp{Mostly skip, but shall we add Hero?: add takens-thiele -- find reference; computability of covering number; mention local PCA based; mention other methods from EGreene report.}

An algorithm for dimension estimation in noise is proposed by \cite{Chen2013}. This algorithm is based on local PCA at multiple scales; here, $\hatdim_{L}$ is the most frequent index of the maximum eigengap of the local covariance matrix. The main challenge is to establish the appropriate range of scales $r$ at which the $d$ principal values of the local covariances separate from the remaining eigenvalues, in noise. The algorithm can be simplified by plugging in the neighborhood radius selected to optimize the Laplacian estimator, by e.g. \citet{Joncas2017-kn}, see Section \ref{sec:epsilon}.

%% file: rev-appli.tex
\subsection{Manifold learning in statistics}

In Section \ref{sec:laplacian} we mentioned that graph Laplacians,
such as $\Lbw$ and $\Lbw^{norm}$ can generate {\em smoothness
  functionals}; given a function $f:\M\rightarrow \rrr$, and its
values on the data points $\bm{f}=[f(\xb_i)]_{i=1}^n$, the value
$\frac{1}{2}\langle\Lbw \bm{f},\bm{f}\rangle$ approximates the $L^2$
norm $\|\nabla f\|^2_{2}$ on the manifold. This can be used as a
regularizer in supervised or semi-supervised learning. If
$\Lbw^{norm}$ is used instead of $\Lbw$, then the smoothness is
measured w.r.t. sampling distribution on $\M$.

Manifold learning by \dmalg~is closely related to {\em spectral
  clustering} \citep{ShiMalik00,MShiaistats01,NJW,ULE,M:spectral-chapter13}, as
both map the data to low dimensions by the eigenvectors of a
Laplacian. For clustering, it is preferable to use $\Lbw^{rw}$ the
random walks Laplacian, which takes into account the data density and
will exaggerate the clusters. In fact, by mapping the data to lower
dimension with $\Lbw^{rw}$, one can observe a continuum between
separating clusters (if the data is clustered) and smooth embedding
(for the data regions where data lie on a manifold), and even perform
simultaneous embedding and clustering. \mmp{figure with aspirin--is
  there space?} In such cases, it is important to calculate a
sufficient number of eigenvectors: for $K$ clusters, there will be
$K-1$ eigenvectors indicating the clustering, and for each cluster,
additional eigenvectors for a low-dimensional mapping of the data in
the respective cluster. If fewer eigenvectors are used, then usually
the clusters will be recovered but not the intrinsic geometry inside
each cluster.

\subsection{Manifold learning for visualization} Embedding algorithms are often used in the sciences for data
visualization. The scientist, as well as the statistician, need to
distinguish between an embedding as defined in Section \ref{sec:bg},
which preserves the geometric and topological data properties, and
other mappings (occasionally also called ``embeddings'') into low
dimensions using embedding algorithms. The latter kind of dimension
reduction is hugely popular, and its value for the sciences cannot be
underestimated. However, the users of dimension reduction for
visualization should be cautioned that the scientific conclusions drawn
from these visualizations must be subject to careful additional
scrutiny, or to a more rigorous statistical and geometric analysis. One pitfall is that when data are mapped into $m=2$ or 3 dimensions,
for visualization, without an estimation of the intrinsic dimension
$d$, the mapping may collapse together regions of the data that are
not close in the original manifold. When clusters are present because
separating the clusters usually requires at least 2 dimensions, most
of the clusters' geometric structure is collapsed. Hence, once
the data is separated into clusters, the cluster structure needs to be
studied by additional dimension reduction. \mmp{figure with digit 3
  here if space} A second pitfall is the presence of artifacts --
interesting geometric features caused by the embedding algorithm but
 not supported by the data. These can be clusters (Figure
\ref{fig:ies}), \hanyuz{filaments \ref{fig:}}, arms, holes or circles, and so
on.

Before assigning scientific meaning to these features, a researcher
should examine whether they are stable, by repeating the embedding
with different initial points, algorithms, and algorithm parameters,
as well as by perturbing or resampling the original data. To assess if
the features are not large distortions, visualizing the distortion
(Figure \ref{fig:chopped_torus_rmetric}) can provide a valuable diagnostic. For example,
when a ``filament'' is produced by stretching a low density region, a
very common effect (see Section \ref{sec:graph}), the
estimated distortion will show the stretching (Figure \ref{fig:chopped_torus_rmetric}),
while for a true filament, the distortion will be moderate \mmp{do we
  have a fig?}. 

\subsection{Manifold learning in the sciences}
\paragraph*{Astronomy and astrophysics} Manifold learning has been used to study data from large astronomical surveys, like the Sloan Digital Sky Survey (SDSS)\footnote{{\tt www.sdss.org}}. The mass distribution in the universe reveals {\em filaments}, i.e. one-dimensional manifolds, and dimension reduction methods, most often Principal Curves (Section \ref{sec:ml} have been used to estimate them \citet{yen-chi}).

Spectra of galaxies are measured in thousands of frequency bands; they contain rich data about galaxies' chemical and physical composition. By embedding these spectra in low dimensions, as in Figure \ref{fig:ies}, one can analyze the main constraints and pathways in the evolution of galaxies \citep{vanderplasConnolly:09}.

\paragraph*{Dynamical systems}
Dynamical systems described by Ordinary or Partial Differential Equations are intimately related to manifolds, while they also exhibit multiscale behavior. Extensions of manifold learning can be used to understand PDE with geometric structure \citep{nadler:06}, study the long term behavior of the system or the ensemble of its solutions \cite{dsilvaTGCoifKev:16,Dsilva2018-dz}.

\paragraph*{Chemistry}
The accurate simulation of atomical and molecular systems plays a major role in modern chemistry. {\em Molecular Dynamics (MD)} simulations from carefully designed, complex quantic models can take millions of computer hours; however, simulations can still be less expensive than conducting experiments, and they return data at a level of detail not achievable in most experiments. Manifold learning is used to discover {\em collective coordinates}, i.e. low dimensional descriptors that approximate well the larger scale behavior of atomic, molecular, and other large particle systems \citep{boninsegnaGobboNoeClementi:15,tribelloCeriottiP:12,noeClementi:17}. In these examples, the systems can be in equilibrium, or evolving in time, and in the latter case, the collective coordinates describe the saddle points in the trajectory, or the folding mechanism of a large molecule \citep{rohrdanzZhengMaggioniClementi:11,dasMSKClementi:06}.

Manifold embedding is also used to create low dimensional maps of families of molecules and materials by the similarity of their properties \citep{ceriottiTParinello:13,isayevFMORTC:15}. 

\paragraph*{Biological sciences}
In neuroscience and the biological sciences, manifold embeddings are widely used to summarize neural recordings \citep{connor.16b,cunninghamYu:14}, to describe cell evolution \citep{Herring2018-cq}\mmp{1-2 more references, with tsne}

\mmp{other tsne applications}

\mmp{yusu wang applications? coifman cell biology}

%% file: rev-conclusion.tex
In practice, ML is overwhelmingly used for visualization
(Section\ref{sec:appli}) and with small data sets. But ML can do much more. Efficient software now exists (\citet{mcqueenMVdpZ:megaman-jmlr16,Policar19},etc) which can embed truly large, high-dimensional data
(for example SDSS). In these cases, ML helps
practitioners understand the data, by e.g. its intrinsic dimension, or
by interpreting the manifold coordinates \citep{samk22:jmlr,boninsegnaGobboNoeClementi:15,vanderplasConnolly:09}. For real data, a manifold learning algorithm has
the effect of smoothing the data and supressing/removing variation orthogonal to the
manifold, which can be regarded as noise, just like in PCA. Finally,
again similarly to PCA, ML can effectively reduce the data to $m\ll D$
dimensions, while preserving features predictive for future
statistical inferences.
Some inferences, such as regression, can be performed on manifold data
without manifold estimation, by for example, local
linear regression \citep{Aswani2011-kd}, or via Gaussian Processes \citep{Borovitskiy20}. A GP on a manifold can
be naturally defined via the Laplacian $\Delta_\M$.

Even when only visualization is desired, care must be taken to ensure
the reproducibility of the results. The implicit assumption that $m=2$ is
sufficient for embedding the data should be validated. Attention to be
paid when the embedding is interpreted: are the features observed
really in the data or artifacts of the algorithm?

\paragraph*{What we omitted}
We surveyed the state-of-the-art knowledge on the main problems and methods of manifold learning, focusing on the algorithms that are proven to recover the manifold structure through learning a smooth embedding.

Among the topics we had to leave out, manifold learning in noise is perhaps the most important one. Noise makes ML significantly more difficult, by introducing biases and slowing the convergence of estimators. This is an active area of research, but the estimation of geometric quantities like tangent space and reach in the presence of noise have been studied (\citet{aamari2018,Aamari2019-dg},etc); the theoretical results of manifold recovery in noise were mentioned in Section \ref{sec:ml}. 

The {\em reach}, or {\em injectivity radius} $\tau(\M)$ of manifold measures how close to itself $\M$ can be. In other words, $\tau(\M)$ is the largest radius a ball can have, so that, for any $\pb\in\M$, if it is tangent to the manifold in $\pb$, it does not intersect $\M$ in any other point. Large $\tau$ implies larger curvature (a subspace has infinite $\tau$) and easier estimation of $\M$~\citep{GenovesePVW12,Fefferman2016-bc,aamari2018,Aamari2019-dg}.
A manifold can have borders; ML with borders is studied for example in \citet{Singer2011VectorDM}, different convergence rates appear when data are sampled close to the border.

Another useful task is embedding a new data point $\xb\in\rrr^D$ onto an existing embedding  $F(\M)$; this is often called {\em Nystrom} embedding~(e.g. \citet{antoine2022}). Conversely, if $\yb\in\rrr^m$ is a new point on the embedding $F(\M)$, obtained e.g. by following a curve in the low dimensional representation of $\M$, how do we map it back to $\rrr^D$? This is usually done by interpolation.

\mmp{removed nn, put in cover-letter}
\mmp{Finally, a word about a related dimension reduction paradigm. In the
neural networks community, it is sometimes stated that internal
representations of neural networks, and more specifically those of
auto-encoders \cite{} are manifolds, representing a possible fourth
paradigm for manifold learning. While deep learning is an entirely
different paradigm for smooth mapping of the data, there are no
guarantees that this mapping has constant rank $d$, even if the
original data lie on a $d$-manifold. Moreover, the intuitions,
mathematics, and techniques for understanding neural networks' internal
representations are entirely different from those surveyed here.}

\mmp{Future of ML?}